\documentclass[runningheads]{llncs}
\usepackage{graphicx}
\usepackage{amsmath,amssymb} % define this before the line numbering.
\usepackage{comment}
\usepackage{color}
\usepackage{tikz}
\usepackage{orcidlink} 
\usepackage{algorithm} 
\usepackage{algpseudocode} 
\usepackage{booktabs}
\usepackage[accsupp]{axessibility}  % Improves PDF readability for those with disabilities.

\begin{document}
% \linenumbers
\pagestyle{headings}
\mainmatter
\def\ECCVSubNumber{29} 

\title{Automatic Analysis of Human Body Representations in Western Art} % Replace with your title

%\begin{comment}
\titlerunning{Automatic Analysis of Human Body Representations in Western Art}
\authorrunning{Zhao et al.}

%\end{comment}
%******************

%\begin{comment}
\titlerunning{Human Body
Representations in Western Art}
\author{Shu Zhao \inst{1}\orcidlink{0000-0001-8129-2583} \and
Alm{\i }la Akda{\u g} Salah\inst{1}\orcidlink{0000-0002-7204-5633} \and
Albert Ali Salah\inst{1,2}\orcidlink{0000-0001-6342-428X}}
\index{Zhao, Shu}
\index{Akda{\u g} Salah, Alm{\i }la}
\index{Salah, Albert Ali}

\authorrunning{S. Zhao et al.}
\institute{Utrecht University, Princetonplein 5, 3584CC, Utrecht, the Netherlands \and
Boğaziçi University, Bebek, 34342, Istanbul, Turkey\\
\email{\{a.a.akdag,a.a.salah\}@uu.nl}}
%\end{comment}
%******************
\maketitle

\begin{abstract}
The way the human body is depicted in classical and modern paintings is relevant for art historical analyses. Each artist has certain themes and concerns, resulting in different poses being used more heavily than others. In this paper, we propose a computer vision pipeline to analyse human pose and representations in paintings, which can be used for specific artists or periods. Specifically, we combine two pose estimation approaches (OpenPose and DensePose, respectively) and introduce methods to deal with occlusion and perspective issues. For normalisation, we map the detected poses and contours to Leonardo da Vinci's Vitruvian Man, the classical depiction of body proportions. We propose a visualisation approach for illustrating the articulation of joints in a set of paintings. Combined with a hierarchical clustering of poses, our approach reveals common and uncommon poses used by artists. Our approach improves over purely skeleton based analyses of human body in paintings. %The pose groups we find automatically are also observed in Warburg's Bilderatlas, which corroborates our findings and shows the usefulness of the approach.
%\dots
\end{abstract}

\section{Introduction}

The human body is expressive of mood and emotions, as well as intentions, and artists have used the expressive possibilities of the body pose to its fullest extent. Body language can reflect embedded societal or gender differences~\cite{noroozi2018survey}, or convey intense emotions, which cannot be discriminated by facial expressions only~\cite{aviezer2012body}. Consequently, art historians analyse the poses of subjects in paintings in depth. 

The portrayal of emotions via poses of a human body was first documented by the cultural and art historian Aby Warburg with his concept of Pathosformel~\cite{impett2016pose}. The term comes from the combination of ``Pathos" (emotion) and ``formel" (a formula). Warburg traces the Pathosformel back to ancient Greek vase paintings~\cite{madhu2020enhancing}, where a narrative is illustrated in the interactions and compositional relationships between characters. The same visual elements of postures and gestures are used in recurrent narratives. Warburg’s \textit{Bilderatlas} contains a rich collection of artifacts through which he studied the adoption of poses by various painters and influences between them. This is one of the reasons why automatic pose estimation in paintings is a useful tool for art historians, as composition transfer can be identified through the similarities between the postures of individual characters~\cite{jenicek2019linking}. Painters often incorporate stylistic elements by copying human poses depicted by other artists. 

The automatic detection and statistical analysis of body shapes and poses in paintings can provide art historians with an overview of artists from a new angle. The statistical analysis in artworks usually focus on genre, style, or artist identification~\cite{wang2020computerized,silva2021automatic}, whereas pose analysis is relatively rare~\cite{impett2016pose,madhu2020enhancing,madhu2020understanding}\footnote{For example IconClass is a classification system for image content and includes human body poses (\url{https://iconclass.org/31A23}).}. It will furthermore be useful in image retrieval for art datasets and archives, as well as for tracking the re-use of visual elements~\cite{castellano2021visual} and help with recent efforts in automatic captioning of paintings~\cite{cetinic2021iconographic,bai2021explain,sheng2019generating}. 

The aim of this paper is to use automatic pose estimation methods to create a tool which will allow pose based analyses of paintings. For this purpose, we start from off-the-shelf pose estimators, and then seek to resolve specific issues pertaining to poses in paintings. More specifically, we combine two pose estimators (to complement each other's shortcomings), followed by an artist-specific normalisation step that corrects occlusions and perspective related distortions based on average body poses depicted by a specific artist. We contribute an artistic pose dataset of Western art that we have semi-automatically annotated for pose ground truth. By applying hierarchical clustering on the detected and corrected poses, we perform a detailed analysis of the joint angles used in paintings in our dataset, and show how the analysis reveals the common and niche pose depictions used and re-used throughout Western art.

This paper is structured as follows. In Section~\ref{section:pose}, we briefly summarise related work on human pose estimation. Section~\ref{section:method} describes our algorithmic pipeline and its various components. Section~\ref{section:data} describes the dataset we have used and annotated in the study. We illustrate our methods via experimental results in Section~\ref{section:experimental}, and conclude in Section~\ref{section:conclusions}.

\section{Related Work}
\label{section:pose}
Human pose estimation from images can be achieved with 2D or 3D models. There are three different types of approaches, using kinematic models (used for 2D / 3D), planar models (used for 2D), and volumetric models (used for 3D). For the kinematic and planar models, body joints are represented with keypoints, and limbs are represented with lines joining those keypoints. While this is suitable for pose analysis in paintings, we will need more than skeleton representations for body depictions, as the same pose can be depicted with different body representations.

We distinguish between top-down and bottom-up approaches for pose estimation. In the top-down pipeline, a human body detector is used to obtain each person's bounding box and a single-person pose estimator is used to predict the locations of keypoints within the bounding box. In the bottom-up pipeline, body joint detectors are used to extract human body joint candidates which are clustered into individual bodies. In general, the bottom-up methods outperform the top-down methods~\cite{zheng2020deep}. OpenPose is one of the most commonly used bottom-up models with a real-time performance to estimate the poses for multiple people in one image~\cite{cao2019openpose}, and we use it to detect keypoints in this work. This is a rapidly growing area, and the keypoint extractor can be updated with more promising approaches. For instance HRNet, while not as widely tested as OpenPose, shows good performance in certain benchmarks~\cite{sun2019deep}. 

Keypoint estimation is not sufficient for representing body shapes in images. DensePose \cite{Guler2018DensePose} is a widely used top-down method, which is built on the Faster R-CNN architecture~\cite{ren2015faster}. In addition to keypoints, it provides a segmentation of body parts (i.e. head, torso, arms, hands, legs and feet) for multiple people in the image. We use this approach in this work for detecting body segments, but further enhance the results with keypoints detected by the more accurate OpenPose.

Pose estimation approaches (as well as face detection), are used for automatic analysis of human representations in paintings~\cite{sari2019automatic,madhu2020enhancing,castellano2021deep}, but also, the analysis of these key-points and landmarks are further used as a way to statistically analyse artistic datasets~\cite{impett2016pose} and to design higher level analysis and synthesis methods~\cite{yaniv2019face,madhu2020understanding,castellano2021visual}. Here we detail three related studies further.

%Computational analysis of artworks using face and body pose analysis is not new~\cite{impett2016pose,madhu2020enhancing,yaniv2019face}. 
Madhu et al. focused on the analysis of figures on Greek vase paintings, which are full of visual narratives, in which the protagonists are depicted by their actions and interactions conveyed through their poses composed against a certain scene~\cite{madhu2020enhancing}. To automatically detect these poses, a styled dataset is generated from the COCO-Persons dataset and a style transfer approach~\cite{huang2017arbitrary} is applied to convert these images to a style similar to Greek vase paintings. Combining a person detector based on Faster R-CNN, and a pose estimator based on HRNet, a model is created and trained on this styled dataset. This model is fine-tuned on a classical archaeology dataset with $2.629$ person annotations and $1.728$ pose annotations from over $1.000$ Greek vase paintings. %The poses are categorised for five different narratives, such as ``Pursuits'', ``Abductions'' and ``Wrestling'' in Agonal and Mythological contexts. The styled model increases the mean accuracy precision (mAP) of pose estimation by $7.62\%$.

In another example by Yaniv et al., facial landmarks are estimated on the portraits~\cite{yaniv2019face}. First, a custom artistic portrait dataset is generated with $160$ paintings from 10 artists. A multi-task cascaded CNN \cite{zhang2016joint} is used to automatically detect the faces, which are then cropped and resized to images of $256 \times 256$ pixels. A landmark detection algorithm is applied to extract $68$ facial landmarks using the Dlib-ml toolkit~\cite{king2009dlib}. With the help of a natural-faces dataset with $68$ landmark annotations per face~\cite{sagonas2013300}, the geometric differences between natural faces and artistic faces is documented. Not surprisingly, artistic faces have a larger geometric variation due to artistic exaggeration and deformation. 

In the third example, a portion of Aby Warburg's Bilderatlas collection is manually annotated via crowdsourcing for pose key-points~\cite{impett2016pose}, and analysed using statistical methods such as hierarchical and two stage clustering to generate an overview of body pose-clusters. To determine the number of clusters automatically, a two-sample Kolmogorov-Smirnov test is run on the distributions
of each joints’ angle. The results show that certain poses from antiquity indeed resurface in Renaissance, however the context of these poses (and hence their emotional content) is changed~\cite{impett2016pose}.

While these studies show the potential of computer vision based analyses in the domain of paintings, annotated datasets are very rare for pose keypoints, body segments or facial landmarks. However, the investment to generate these datasets is worthwhile.  One possibility is to apply style
transfer to already-annotated datasets of photographic images, and use such
an augmented dataset as a training set for pose and landmark estimation in
paintings. However, paintings may contain more stylised poses than naturally
occurring poses in photographs, and manual annotation of paintings can potentially result in better model training. In the end, both approaches are costly, either in terms of computing or manpower. In this paper, we explore to what extent automatic pose estimation with widely used off-the-shelf tools can be used for body pose and shape analysis in paintings.

\section{Methodology}
\label{section:method}
\begin{figure}
  \centering
  \includegraphics[width=1.0\linewidth]{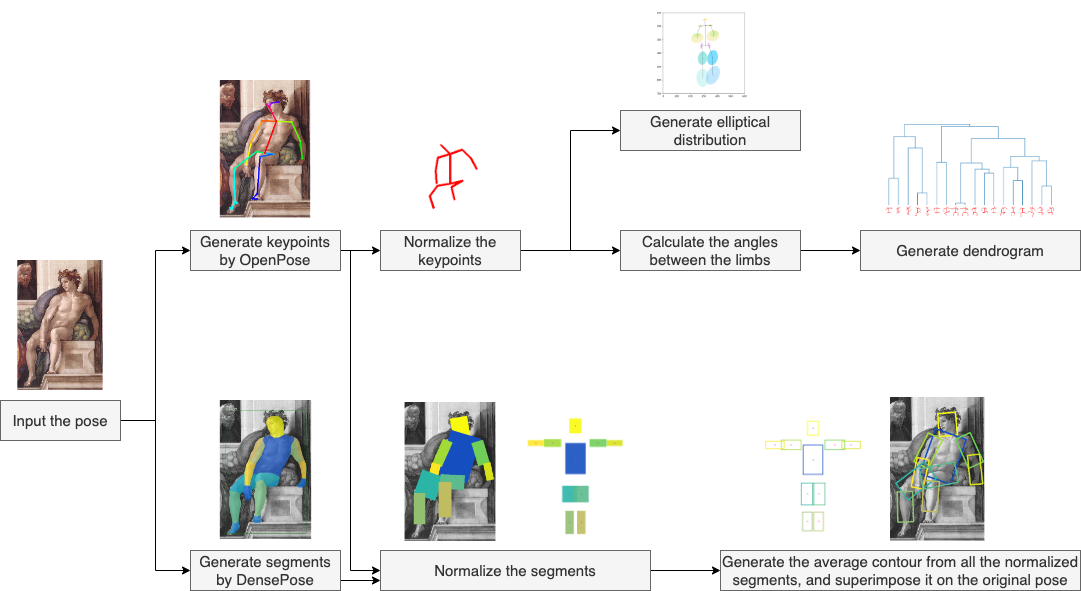}
  \caption{\label{fig:method-pipeline}
           The pipeline of pose analysis.}
\end{figure}

Our general analysis approach is illustrated in Figure~\ref{fig:method-pipeline}. We use OpenPose and DensePose in parallel to obtain a set of keypoints and segments for the depicted bodies in the paintings. For pose analysis of a group of paintings (such as from a single painter or a style), we use the keypoints from OpenPose, prepare visualisations that depict distributions of joint angles, and following~\cite{impett2016pose}, dendrograms to find pose groups. For body shape analysis, we use a shape normalisation step, and generate the average contours from all normalised segments to superpose them on original poses.

\subsection{Average contours of body segments}
In Western art historical literature, the archetype, or the canon is an example that is considered as the highest standard that transcends the aesthetics of a given era~\cite{langfeld2018canon}. Such an example can be an artwork, but also the depiction of a pose, or more importantly, the human body. The canonical body, i.e. the perfect proportions of a the human body has been a topic of discussion and elaboration since the ancient Greeks~\cite{murtinho2015leonardo}. There have been many descriptions of these proportions, as well as drawings. Among these the Vitruvian Man by Leonardo da Vinci is one of the most well known~\cite{magazu2019vitruvian}.

In our approach, we use the Vitruvian Man as a standard for normalising extracted body contours of artistic poses, as it has been influential in defining the ideal body proportions in Western art. This allows us to compare different poses with each other at the same scale. Using contours, we propose an approach to mitigate the occlusion and perspective issues that are generally present in automatic pose estimation, as well as generate an overview for each artist's preferred style in drawing human bodies: Do they use proportions closer to norm, or do they prefer exaggeration of these by elongating or thickening of body segments? Lastly, through the comparison of average contours we illustrate the differences between the contours of natural poses and artistic poses. %With this exercise we further want to test if the abstracted contours of a pose can to some extent mitigate the occlusion and perspective issues that are generally present in 2D DensePose inference. The exercise furthermore allow us to generate an overview for each artist's preferred style in drawing human bodies: Do they use proportions closer to norm, or do they prefer exaggeration of these by elongating or thickening of body segments?  

To carry out the normalisation process we follow several assumptions: 1) The head size is different for men and women. The (vertical) size of the head is equal to the vertical distance between the nose and the top of the chest. 2) The body proportions are the same for men and women; 3) The body segments are convex, and can be abstracted as rectangles; 4) The body segments are symmetrical between left and right sides. A specific segment is also symmetrical around its centroid.

We use the head size as a normalisation reference, based on which we scale the length of limbs and area of segments. The baseline Vitruvian Man is referenced by a fixed dimension, i.e., $\left(624 \times 624 \right)$ in pixels, with male head size being fixed at $\textbf{62}$ pixels, and the female head size by $\textbf{58}$ pixels. The median size, i.e. $23.2$ centimetres for men and $21.8$ centimetres for women, is taken as the norm\footnote{\url{https://en.wikipedia.org/wiki/Human\_head}}. We transform this proportion to $\textbf{62}$ pixels for men, and $\textbf{58}$ pixels for women. 
%\begin{equation} \label{eqn:scale-factor-genders}
%    \frac{Standard \ head \ height \ of \ women}{Standard \ head \ height \ of \ men} = \frac{21.8}{23.2}
%\end{equation}
For an input pose from any image, the scale factor is calculated by dividing the standard head size to the actual head size in pixels. Other limbs are scaled accordingly by the same scale factor. %Thus, all poses whose measuring unit is in pixels can be compared relatively.

%\begin{equation} \label{eqn:scale-factor-input}
%    Scale \ factor = \frac{Standard \ head \ height \ in \ pixels}{Actual \ head \ height \ in \ pixels}
%\end{equation}

%To scale the limbs that are represented by the distance between keypoints, and to scale the segments, i.e., areas rather than lines, we use centroids of segments as the pivot around which to scale surrounding pixels within the boundary of corresponding segments. To demonstrate this by an example, the centroid of the head is the average point of an area occupied by the head segment, which is later mapped on the Vitruvian Man as the midpoint between ``below the chin'' and ``the top of the head''. %The centroid of torso is the midpoint between the keypoints neck and midhip, which is mapped on the Vitruvian Man also as the midpoint between the neck and the midhip. The centroids of arms and legs are the midpoints between the keypoints wrist and elbow, elbow and armpit, hip and knee, knee and ankle, which are mapped on the Vitruvian Man as the midpoints of the corresponding joints. All the mapped centroids are shown in \autoref{fig:vitruve-canon} as white dots. To differentiate, the centroid can be either an average point of an area, or a midpoint between two keypoints, around which each segment can be scaled area-wise during normalisation. On the other hand, the midpoint on the Vitruvian Man is a fixed point on the standard image, to which each normalised segment can be translated, so that the centroid of each segment can be superimposed on the midpoint of the Vitruvian Man.
The output of normalisation is a T-pose figure superimposed on the Vitruvian Man with $10$ segments, i.e., head, torso, upper and lower arms, upper and lower legs, respectively. We use the keypoints detected by OpenPose and match these to the DensePose segments, as OpenPose has a higher  accuracy for keypoint detection (see \autoref{tab:accuracy-keypoints}). If a painting has multiple interacting poses, bounding boxes of segments may overlap. In such cases we will discard the data during matching. We also discard instances if the torso is only partially detected since we use this information for rotating the whole pose to a vertical position. We furthermore rotate each segment separately to a vertical or a horizontal position, as required by a T-pose. Lastly we dilate all the segments to their tightest-fitting rectangles. %The pose  normalisation method is summarised in Algorithm \autoref{alg:normalisation}. 

\subsection{Visualising joint distributions}
To summarise artists style when it comes to pose geometry, we prepare and visualise distributions for male and female poses by using angle of joints as a reference point. This type of analysis gives on overview of the range of poses for a group of paintings (examples shown in Section~\ref{subsection:ellipsoids}).

The procedure starts with the normalisation of keypoints following these three steps: 1) Validate whether a set of indicative keypoints (we use nose, neck and midhip), are detected. 2) Rotate the whole pose to a vertical position, so that the spine is vertical. 3) Use $60$ pixels (size of the head) as a scaling factor to normalise all the poses. 

After this procedure is applied to all poses in the selected set, the distributions of the keypoints are visualised by fitting Gaussian ellipsoids on the keypoint locations. Each ellipsoid can be thought of a visualisation of the covariance matrix of the spread of the points. Hence, each ellipsoid stands for the standard distribution of one keypoint for a group of poses. However, while traditionally one standard deviation is used for the contours, we use here half a standard deviation for better visibility. The covariance matrix $\sum$ of the vector for one keypoint $X=[X1,X2]$ can be calculated and represented as $
\sum = 
\begin{bmatrix}
\sigma_{11} & \sigma_{12} \\
\sigma_{21} & \sigma_{22}
\end{bmatrix}$.
%\end{equation}

The coordinate $(x, y)$ of the ellipsoid for the keypoint with its corresponding radius and scale around it is as follows:
%\begin{equation}
\begin{align*}
x &= \frac{\sum_{i=1}^{n} X1_i}{n} & radius_{x} &= \sqrt{1 + \frac{\sigma_{12}}{\sqrt{\sigma_{11} \times \sigma_{22}}}} & scale_{x} = \sqrt{\sigma_{11}} \times 0.5 \\
y &= \frac{\sum_{i=1}^{n} X2_i}{n} & radius_{y} &= \sqrt{1 - \frac{\sigma_{12}}{\sqrt{\sigma_{11} \times \sigma_{22}}}} & scale_{y} = \sqrt{\sigma_{22}} \times 0.5 \\
\end{align*}
%\end{equation}

\subsection{Hierarchical clustering}
The visualisation of joint distributions explains the range of joints depicted in the poses, whereas by looking at joints in relation to each other it is possible to understand the nature of poses as well. To do that, we carry out hierarchical clustering for artistic poses~\cite{impett2016pose}. This process can also illustrate the common and niche poses in detail, which can to some extent explain why a pose detector performs better or worse for some poses. 

For hierarchical clustering, the keypoints are normalised in three steps: 1) Validate whether all the $6$ torso keypoints are detected, i.e., neck, right and left shoulders, midhip, right and left hips. 2) Rotate the whole pose to a standing-up position, with the spine being vertical. 3) Calculate the inner angles for all the triplets of joints. In total, there are $13$ such joint triplets, i.e., $1$ triplet of (nose, neck, midhip), $6$ triplets of the right body: (shoulder, neck, midhip), (elbow, shoulder, neck), (wrist, elbow, shoulder), (hip, midhip, neck), (knee, hip, midhip), (ankle, knee, hip), and $6$ symmetric triplets of the left body. The inner angles of each pose on the 2-dimensional plane are treated as a $13$-dimensional vector, representative of each pose. These pose vectors are used for agglomerative hierarchical clustering, in which each pose starts in its own cluster, which is merged with other clusters if their pair-wise Euclidean distance is the smallest. The results are depicted in a dendrogram (see Section~\ref{subsection:hierarchical}), where similar poses are connected with each other by shorter distances, whereas different pose clusters are connected with each other by longer distance. 
\section{The Artistic Pose (AP) Dataset}
\label{section:data}
%There are three tasks that we address in this paper: (1) testing off-the-shelf pose estimation methods on Western artworks; (2) using statistical analysis to improve pose estimation results; and (3) extracting and analysing the geometric shapes of artistic poses for a selected set of artists. To complete these tasks successfully, we need well-annotated datasets, where the joints and segments of the human bodies are either manually marked and masked, or correctly inferred by trained pose estimation models.

%As mentioned in Section~\ref{section:pose}, there are only two artistic-pose datasets that are already manually annotated. These datasets do not focus on individual artists, hence a geometrical analysis of poses drawn by a given artist is not possible with these datasets. 

%Here, the challenge is the trade-off between accuracy and effort. The highest accuracy results from the manual annotation of the poses from selected paintings, but this needs the most effort. The second option is to rely on the inferred outcome of OpenPose and DensePose, but the question is whether the inferred joints and body segments are accurate enough to base pose analysis upon.

To test the inference accuracy of OpenPose and DensePose we decided to generate a dataset by selecting 10 artists from the {Painter by Numbers} dataset which has $103,250$ paintings from Western art that range from the early 11th century to the 2010s. This is similar in size to earlier studies in the literature~\cite{yaniv2019face}. We chose artists with more nude paintings, as these were expected to generate the least problems for pose estimation, but the earlier painters have less nude paintings.
We also tried to find a gender balance, as well as a genre balance in choosing our artists. 
% did you find male nudes as well?
The resulting Artistic Pose dataset (from now on called the AP Dataset\footnote{The dataset with manual annotations, as well as all the code are made publicly available at \url{https://github.com/tintinrevient/joints-data}.}) covers a wide range both in time and style: Michelangelo (Renaissance), El Greco (Mannerism), Artemisia Gentileschi (Baroque), Pierre-Paul Prud'hon (Romantism), Pierre-Auguste Renoir (Early Impressionism), Paul Gauguin (Post Impressionism), Felix Vallotton (Magical Realism) and Amedeo Modigliani (Expressionism), Tamara de Lempicka (Art Deco), and Paul Delvaux (Surrealism).
% AAS: 
%``\href{https://www.kaggle.com/c/painter-by-numbers/data} add this to bib

We furthermore make use of a natural-pose dataset to bring forth an understanding of the geometric style of natural poses and to compare them to the artistic poses. {COCO Persons dataset} is one of these datasets with a wide variety of common activities that are manually annotated with respect to the joints and body segments~\cite{lin2014microsoft}. More importantly, OpenPose and DensePose are trained with this dataset. 
%\add this to bib: href{https://cocodataset.org}

\section{Experimental Results}
\label{section:experimental}

\subsection{Comparison of OpenPose and DensePose}

We first apply OpenPose and DensePose on the AP dataset and evaluate their performances.
To measure accuracy of keypoint detection, we use PCK (Percentage of Correct Keypoints). The keypoints are considered correct if the distance between the inferred and the true keypoint is within a certain threshold. PCK for each artist is shown in Table~\ref{tab:accuracy-keypoints}. For pose analysis, we only use $15$ keypoints, namely the nose, neck, midhip, left and right shoulders, elbows, and wrists, left and right hips, knees, and ankles, respectively. %For all artists, the keypoints are solely inferred by OpenPose and DensePose, and a manual evaluation has been carried out.  

On average for all artists, the inference accuracy of OpenPose is around $80\%$, and $66\%$ for DensePose. The inference of joints by the paintings of Tamara de Lempicka and Amedeo Modigliani have the least accuracy due to their exaggeration of shapes and proportions. In contrast, Paul Delvaux's paintings have the highest accuracy by both OpenPose and DensePose, because most of the poses are resting or standing with naturally hanging arms, which constitute easy cases for prediction.

\begin{table}
\begin{center}
\caption{The measurement of PCK. The columns list the artists, the total number of people depicted (Male and Female), the number of valid keypoints and segments, accuracy of DensePose (Acc.), PCK of OpenPose (PCK-O) and DensePose (PCK-D)}
\begin{tabular}{lccclll} 
\toprule
Artist & Subjects & \# Keypoints & \# Segments & Acc. & PCK-O & PCK-D \\ 
\midrule
Michelangelo & 15 (14M, 1F) & 214 & 163 & 79\% & 87\% & 63\%  \\

El Greco & 34 (32M, 2F) & 375 & 259 & 42\%& 73\% & 55\% \\

Artemisia Gentileschi & 21 (6M, 15F) & 214 & 182 & 59\% & 86\% & 60\%\\

Pierre-Paul Prud'hon & 15 (7M, 8F) & 216 & 196 & 65\% & 93\% & 82\% \\

Pierre-Auguste Renoir & 19 (19F) & 237 & 200 & 49\%& 80\% & 63\%  \\

Paul Gauguin & 31 (4M, 27F) & 414 & 335  & 64\% & 82\% & 74\%\\

Felix Vallotton & 20 (1M, 19F) & 243 & 206 & 66\% & 86\% & 67\% \\

Amedeo Modigliani & 15 (15F) & 180 & 147 & 21\%& 40\% & 30\% \\

Tamara de Lempicka & 18 (1M, 17F) & 227 & 188 & 54\%& 69\% & 52\% \\

Paul Delvaux & 35 (4M, 31F) & 455 & 371 & 82\% & 89\% & 89\% \\
\bottomrule
\end{tabular}
%\vspace{5pt}
\label{tab:accuracy-keypoints}
\end{center}
\end{table}

\begin{figure}[htb]
  \begin{center}
  \resizebox{.99\textwidth}{!}{
  \includegraphics[height=2cm]{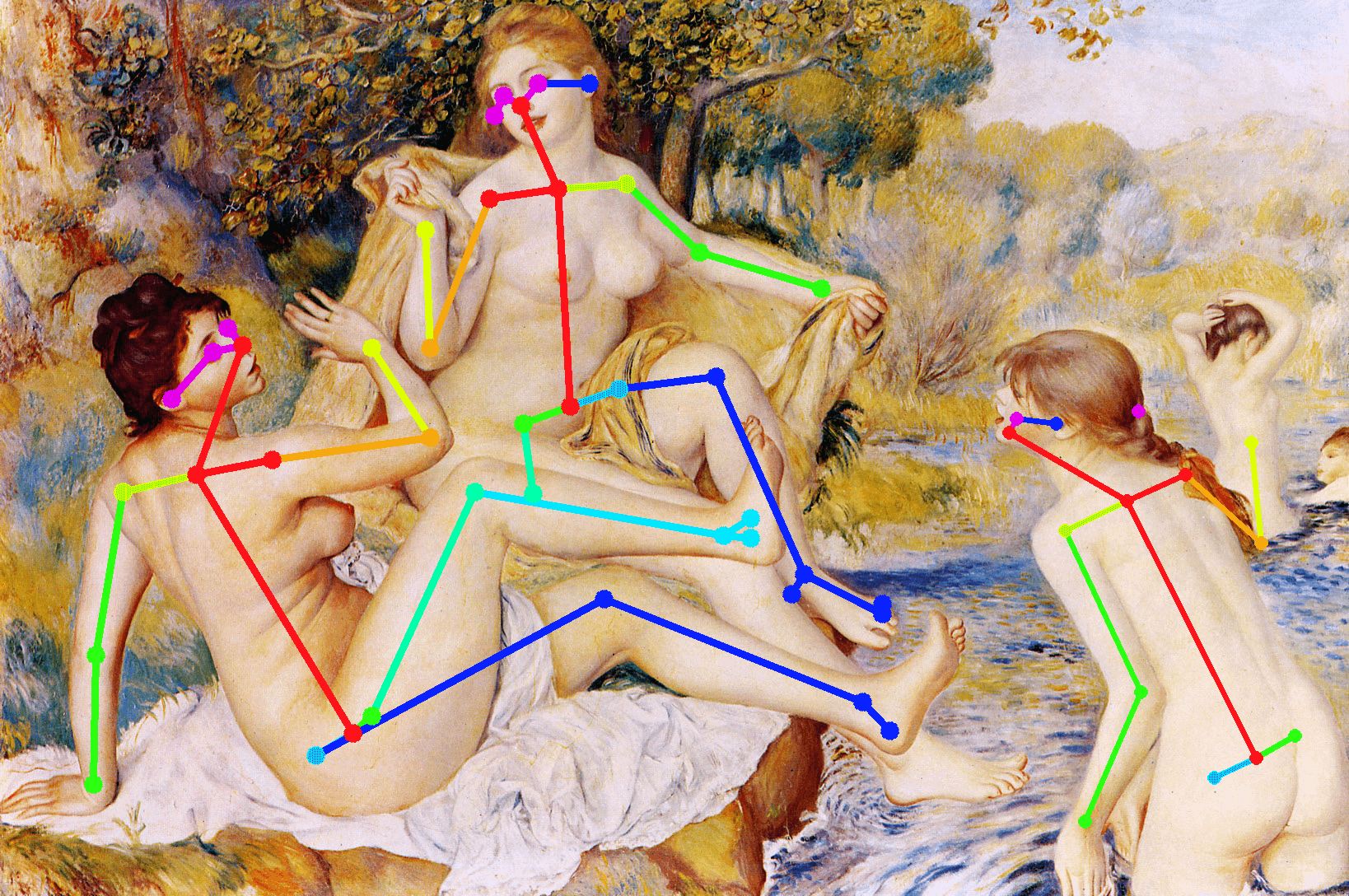}
  \includegraphics[height=2cm]{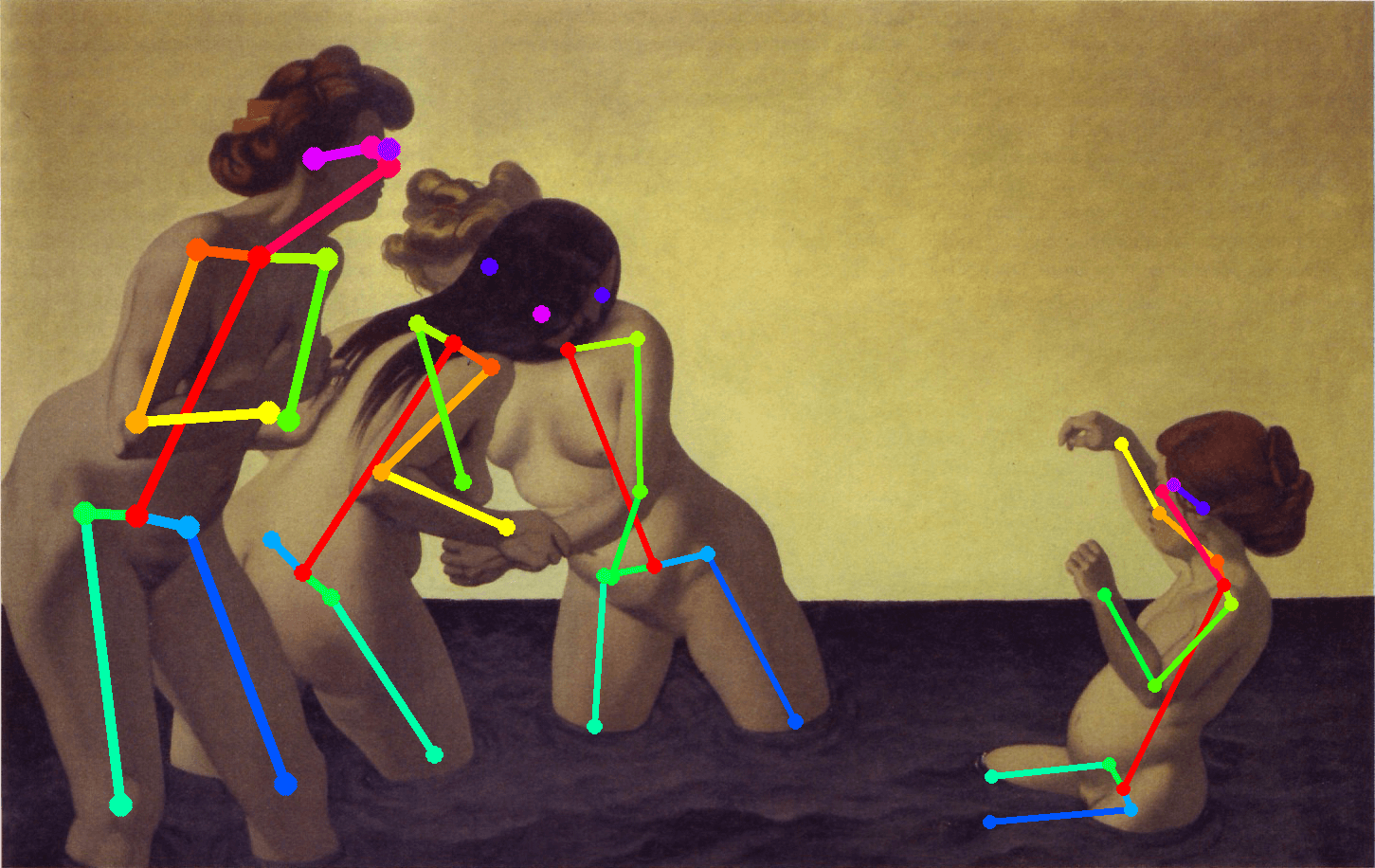}
  \includegraphics[height=2cm]{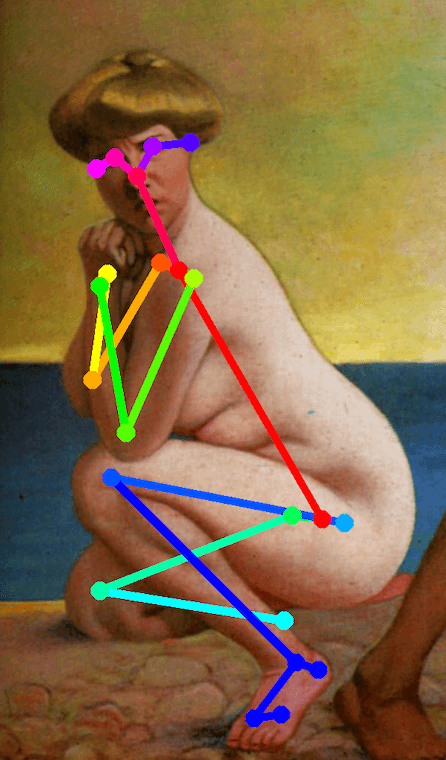}
  \includegraphics[height=2cm]{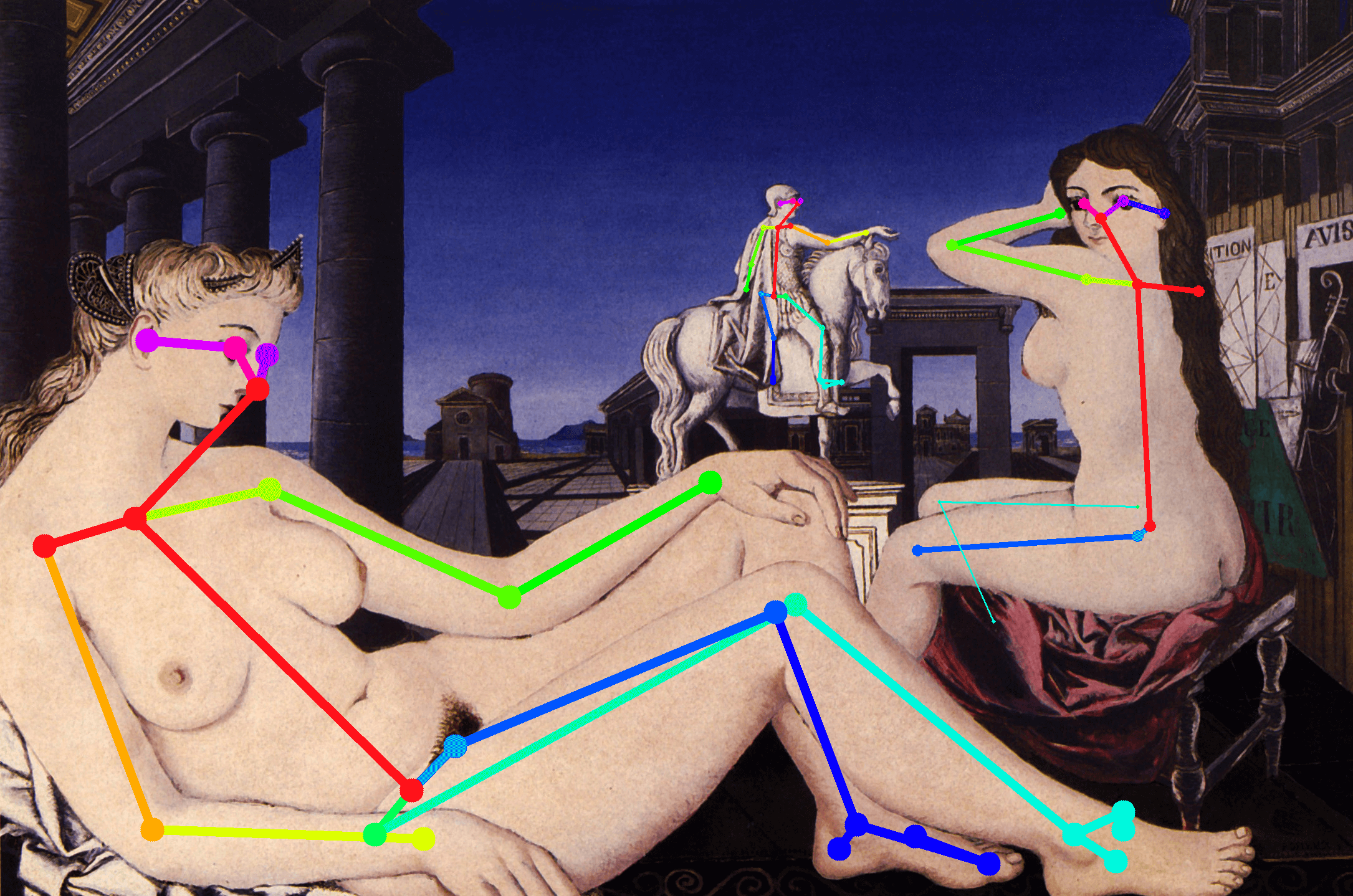}
  }

  \resizebox{.99\textwidth}{!}{
  \includegraphics[height=2cm]{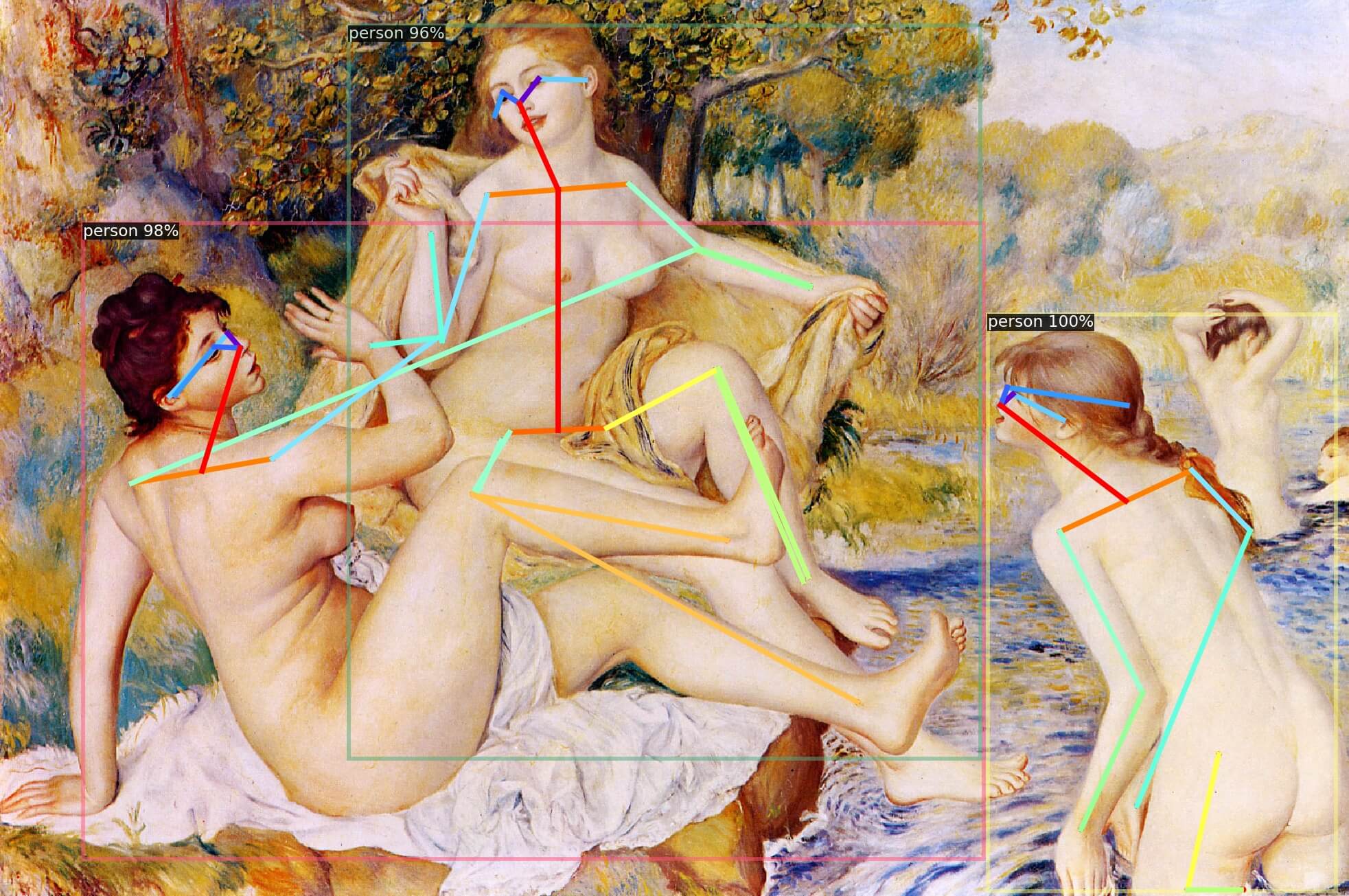}
  \includegraphics[height=2cm]{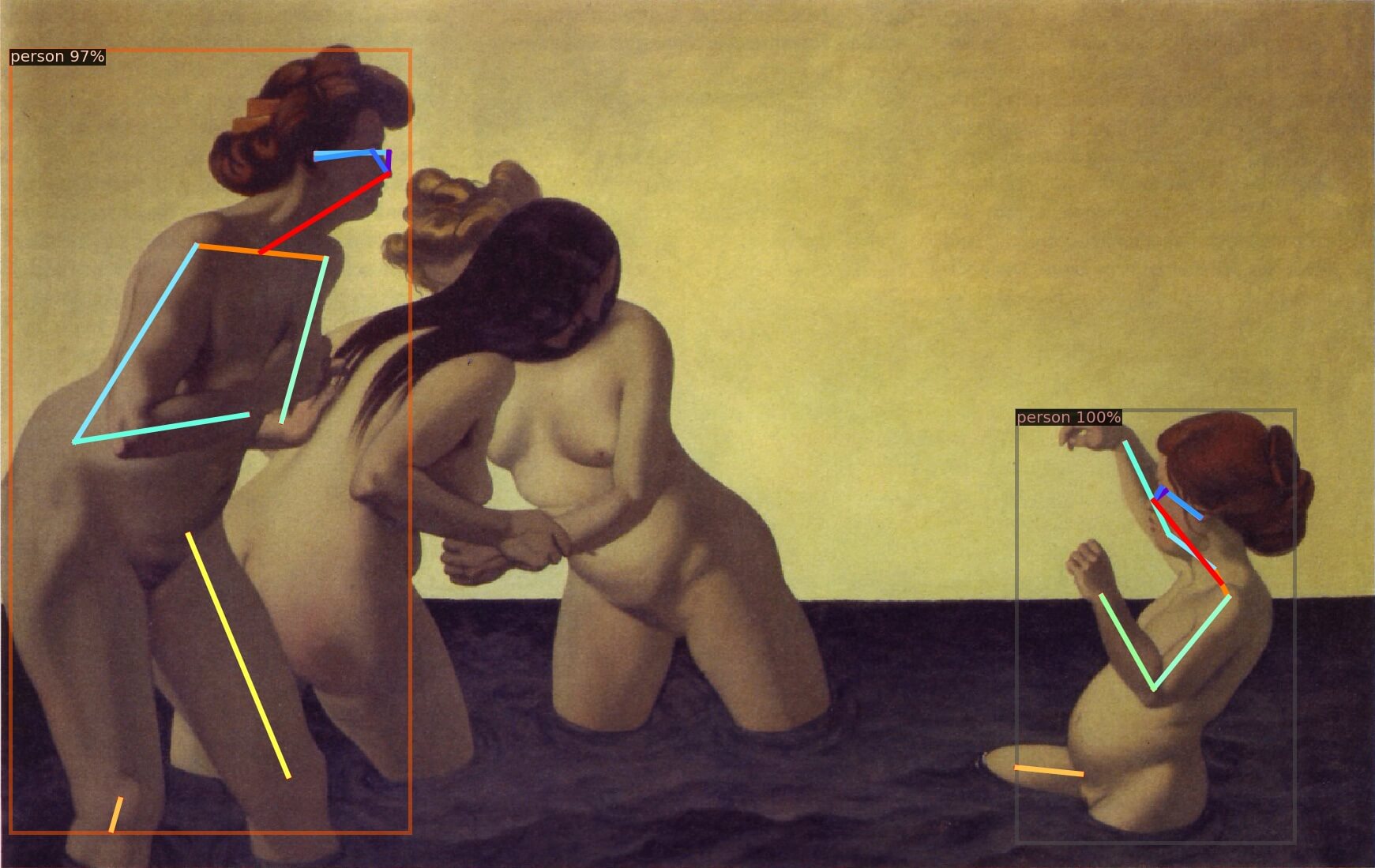}
  \includegraphics[height=2cm]{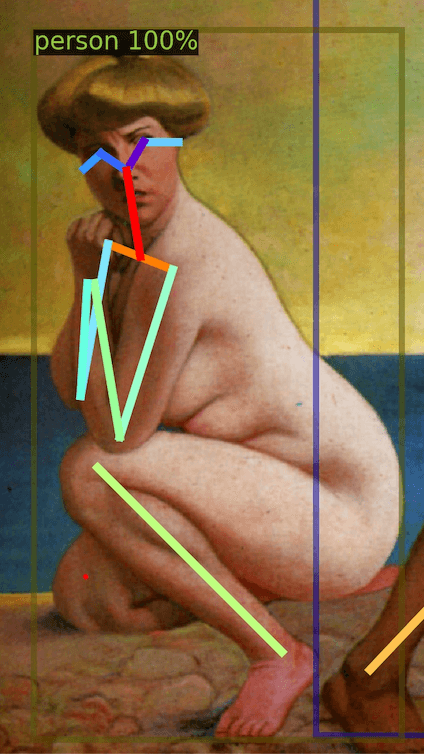}
  \includegraphics[height=2cm]{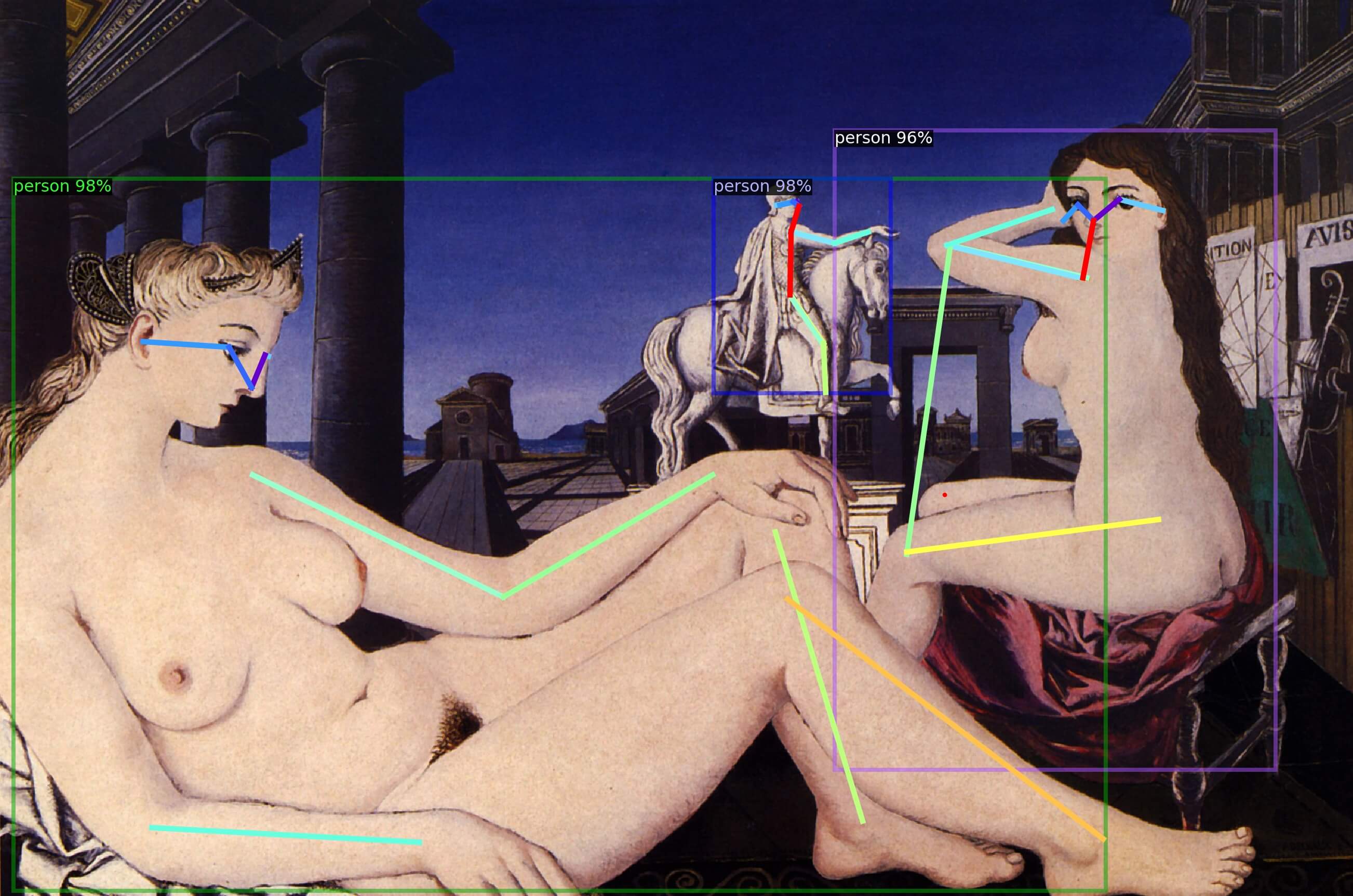}
  }

  \resizebox{.99\textwidth}{!}{
  \includegraphics[height=2cm]{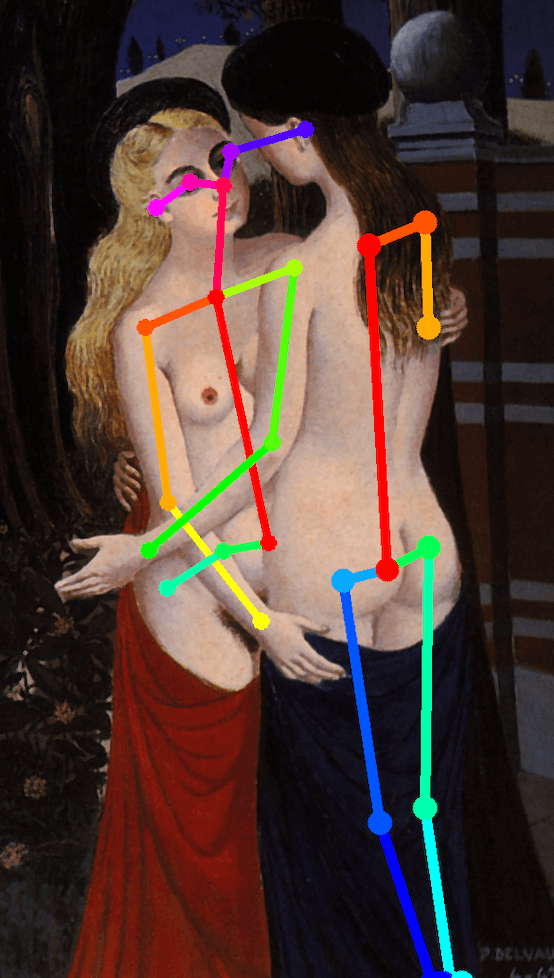}
  \includegraphics[height=2cm]{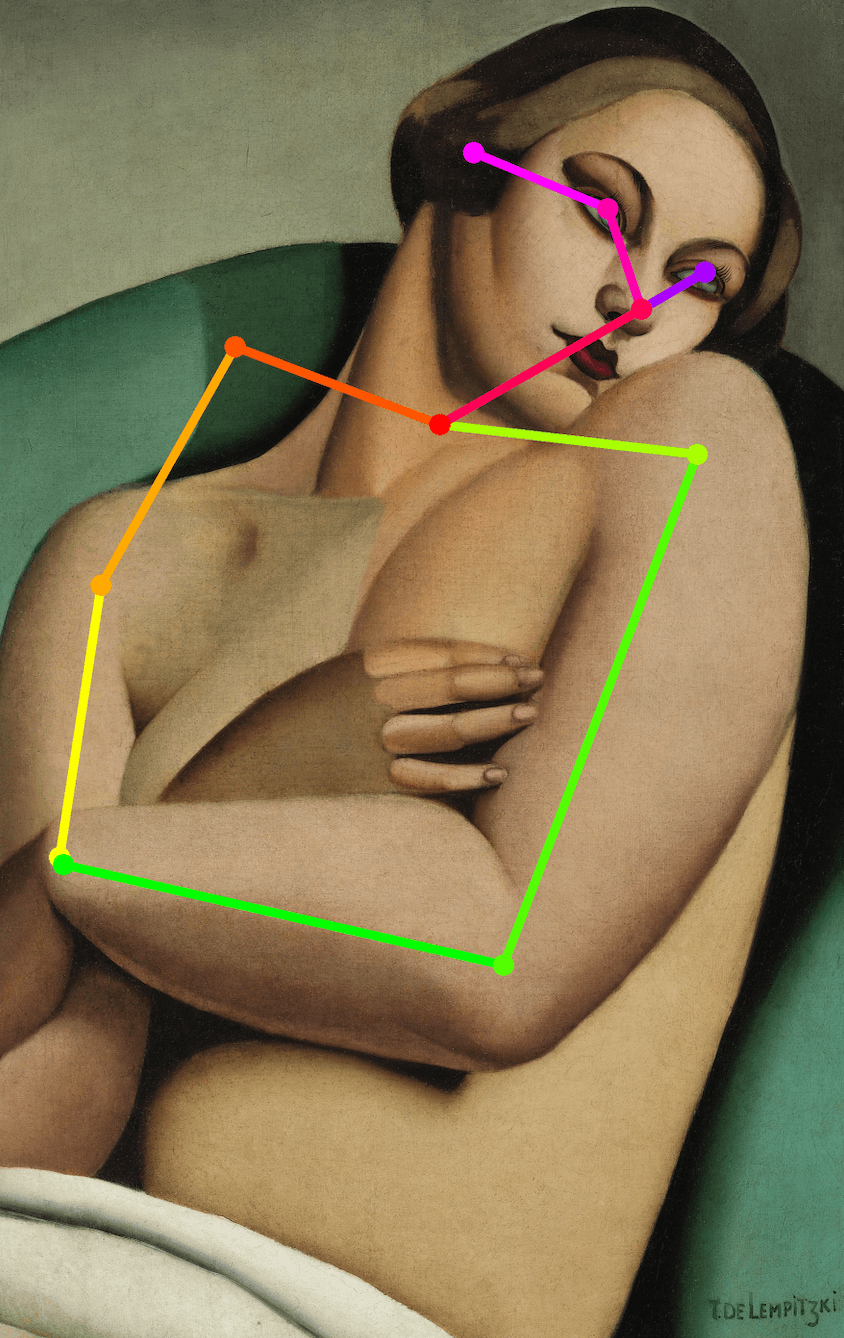}
  \includegraphics[height=2cm]{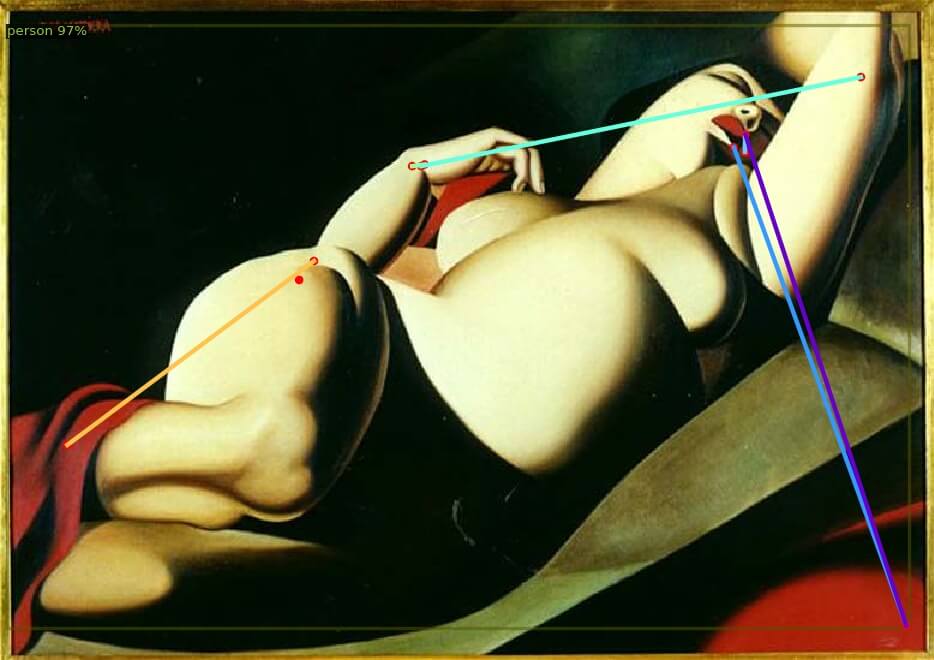}
  \includegraphics[height=2cm]{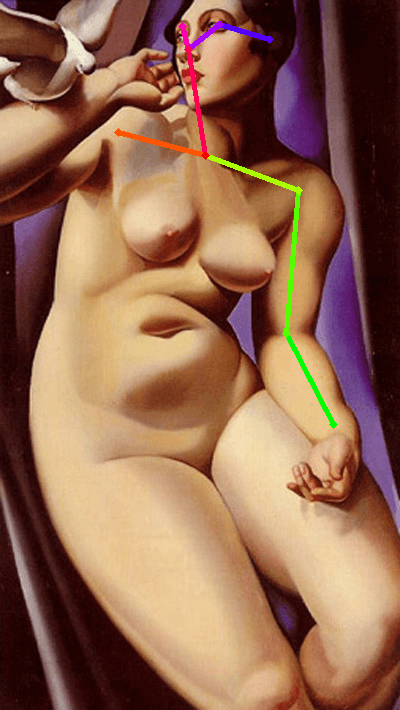}
  \includegraphics[height=2cm]{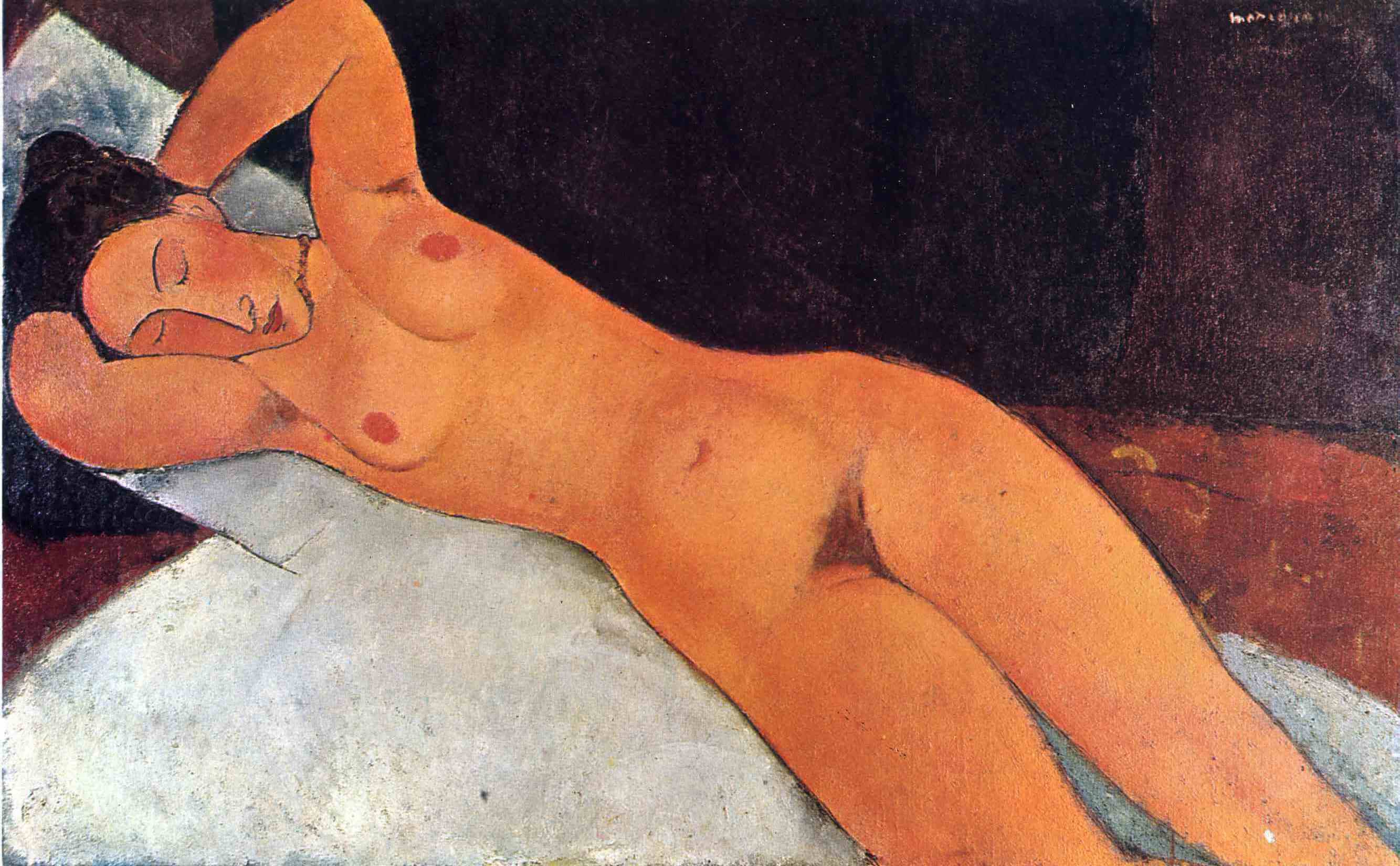}
  }
  \end{center}
  \caption{\label{fig:openpose-densepose}
           The joints inferred by OpenPose (first row) and DensePose (second row) under scenarios such as multiple people, low contrast, and niche poses. The last row shows the more difficult scenarios for both OpenPose and DensePose.}
  
\end{figure}

In general, OpenPose outperforms DensePose, especially under scenarios such as multiple people, low contrast, and niche poses, as shown in Figure~\ref{fig:openpose-densepose}. The universally difficult scenarios are when (1) two people hug or interact closely with each other, (2) the poses have twisted limbs, (3) there are niche perspectives, (4) there are exaggerated shapes, (5) there are exaggerated body proportions. These challenging scenarios are illustrated in the last row of Figure~\ref{fig:openpose-densepose}, in the mentioned order. 

Next, we test the accuracy of DensePose body segment detection, by the percentage of correctly inferred segments over the total number of visible segments (max. 14), as shown in the last column of Table~\ref{tab:accuracy-keypoints}. The factors that impact detection are the number of people in the painting, the occlusions of clothes, interacting people, niche poses and perspectives, artistic effects including unusual colouring and brush usage for body segments, and unusual shapes of body segments.

\begin{figure}[htb]
  \begin{center}
  \resizebox{.99\textwidth}{!}{
  \includegraphics[height=3cm]{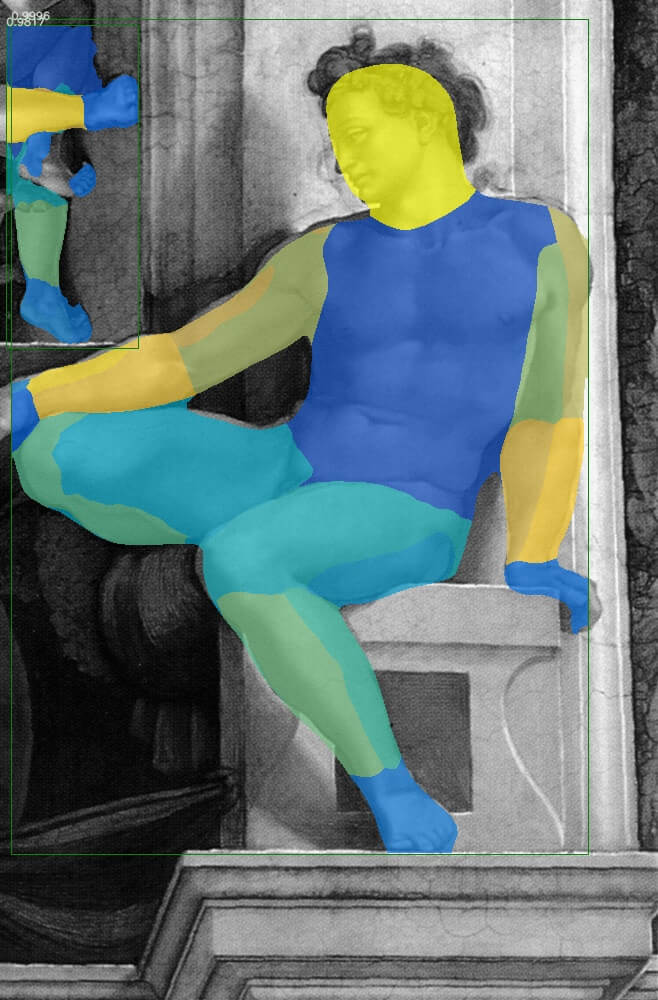}
  \includegraphics[height=3cm]{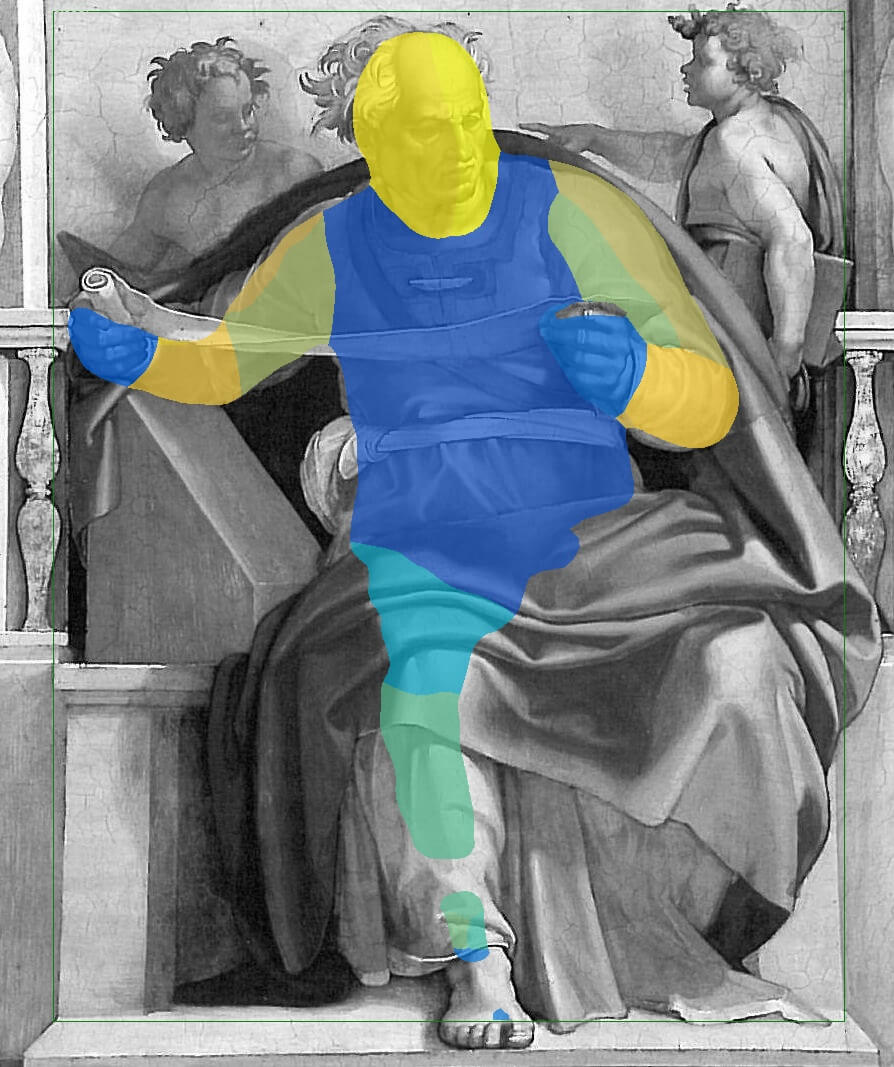}
  \includegraphics[height=3cm]{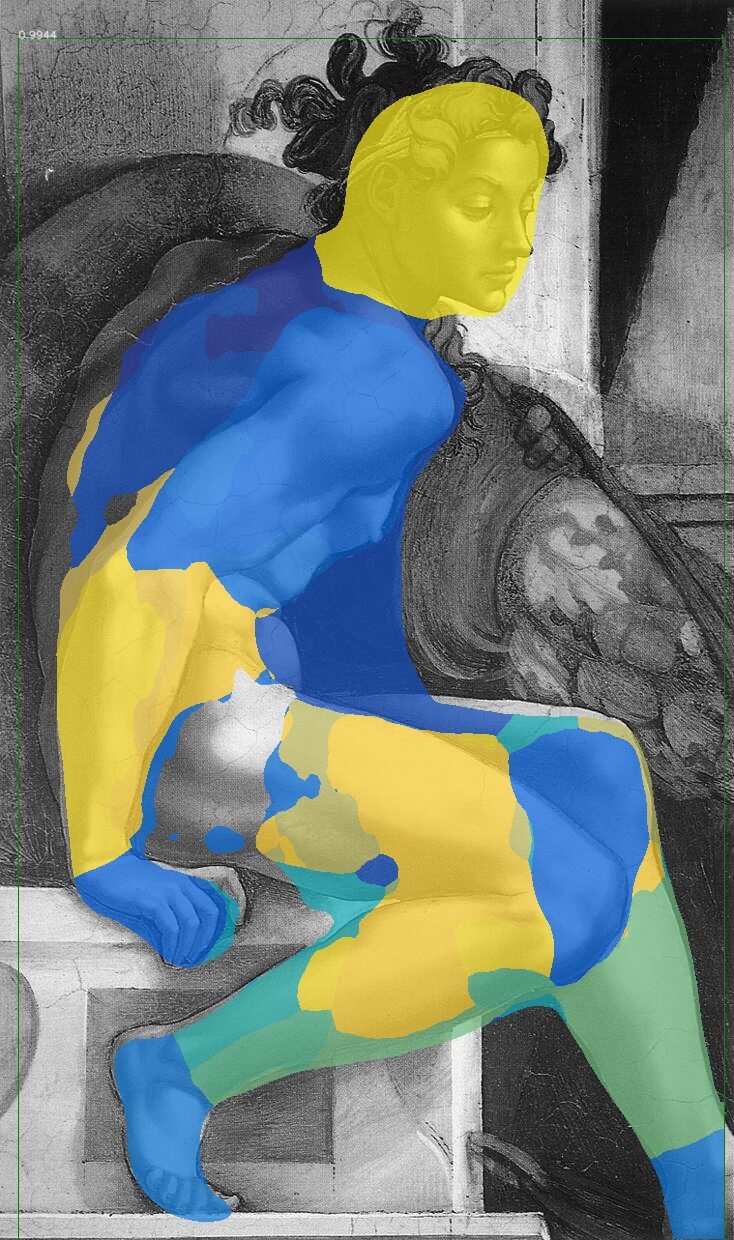}
  \includegraphics[height=3cm]{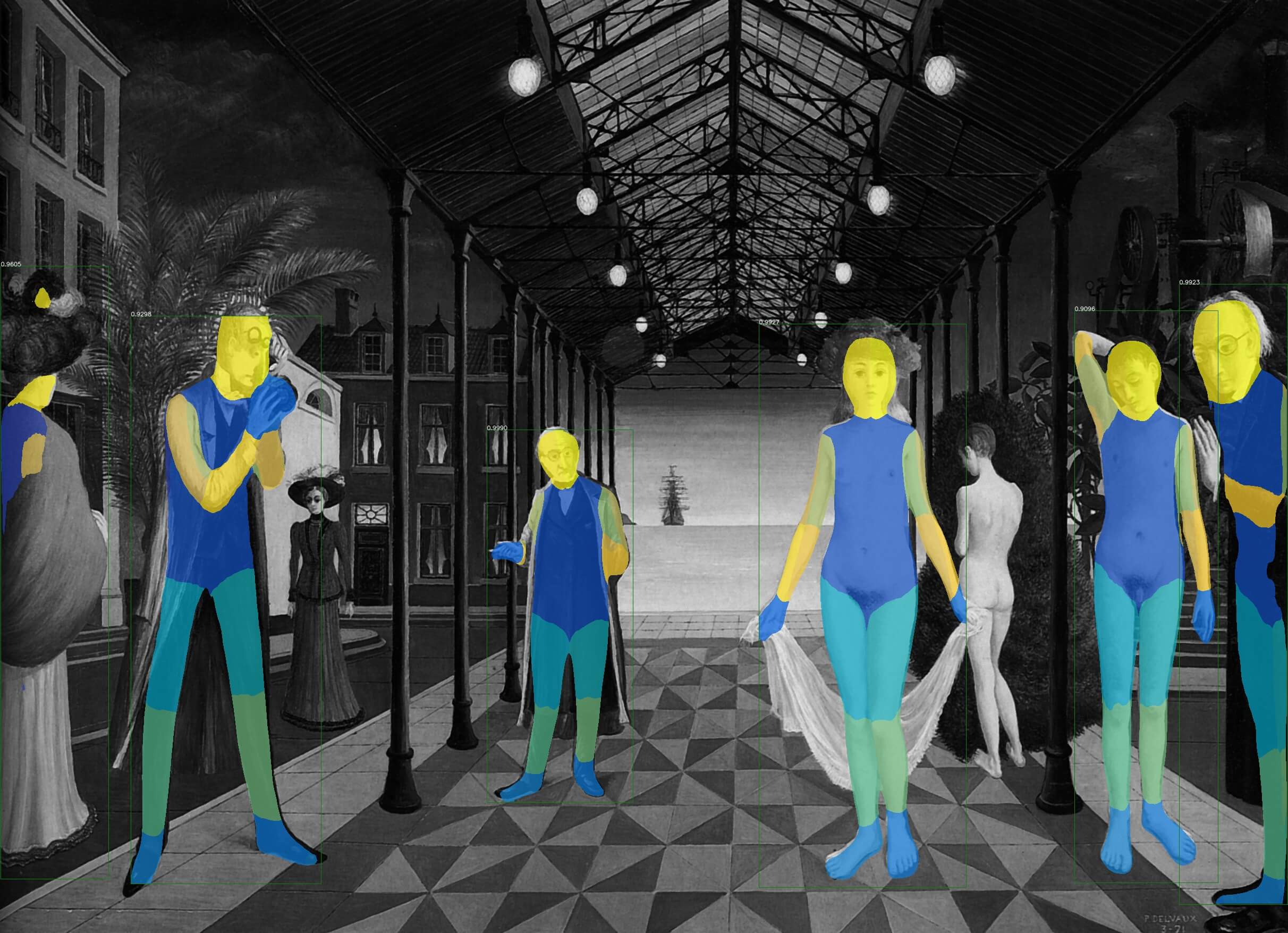}
  \includegraphics[height=3cm]{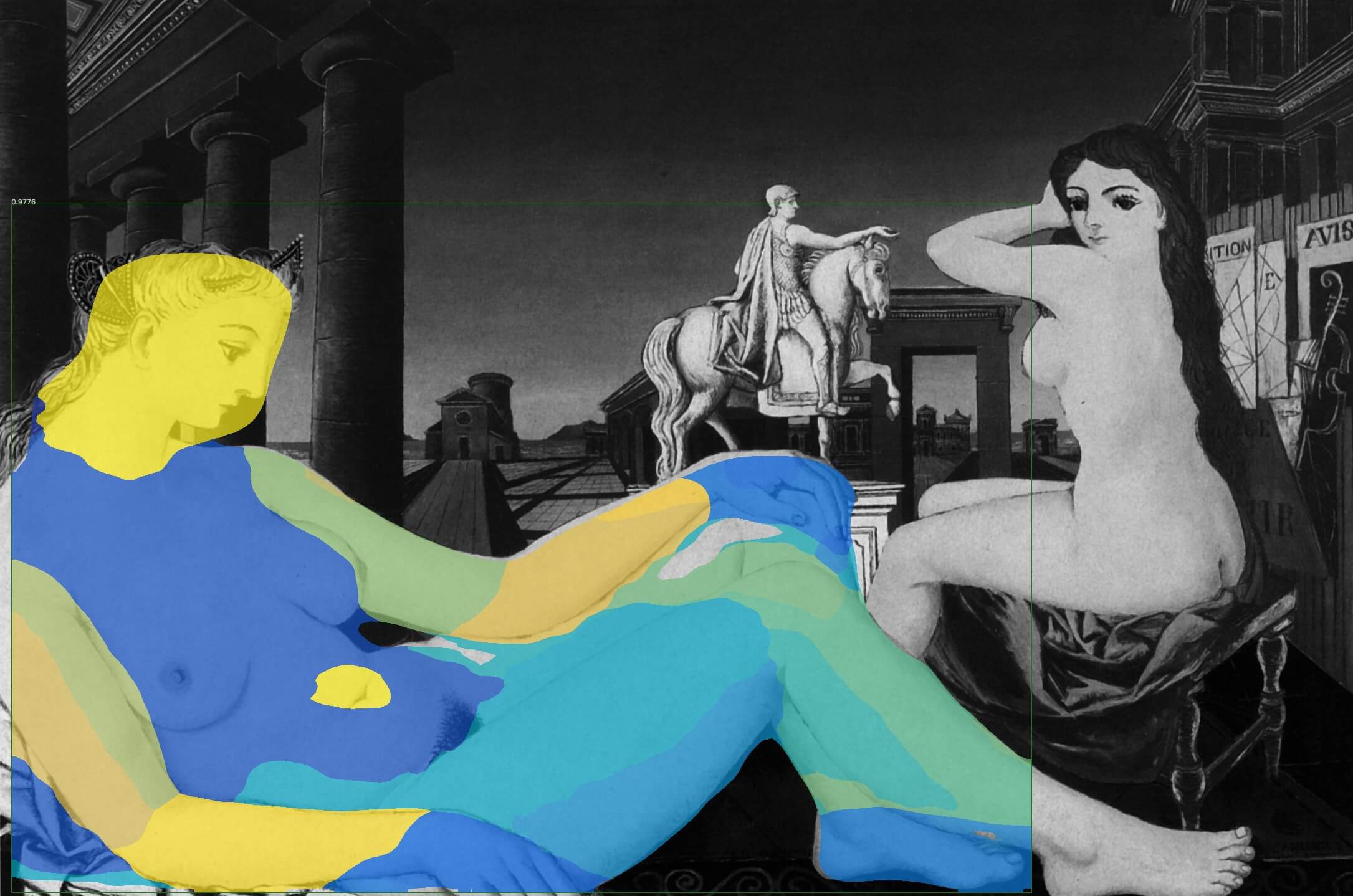}
  }
  
  \resizebox{.99\textwidth}{!}{
  \includegraphics[height=3cm]{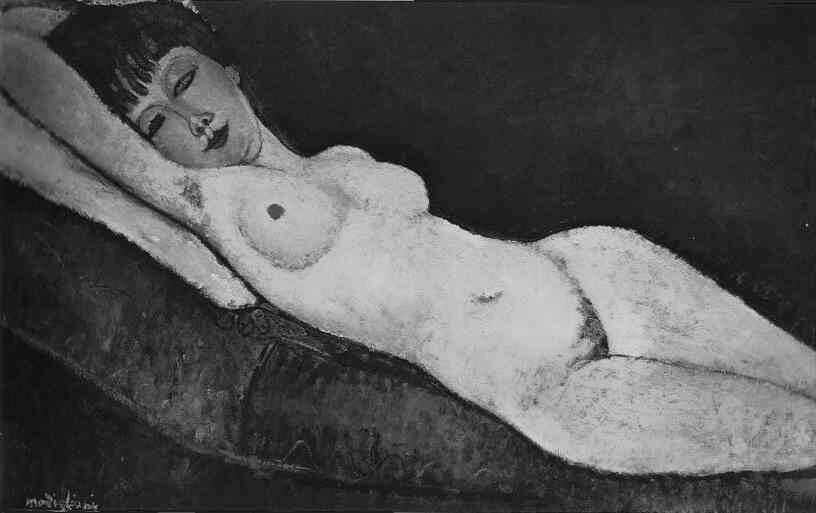}
  \includegraphics[height=3cm]{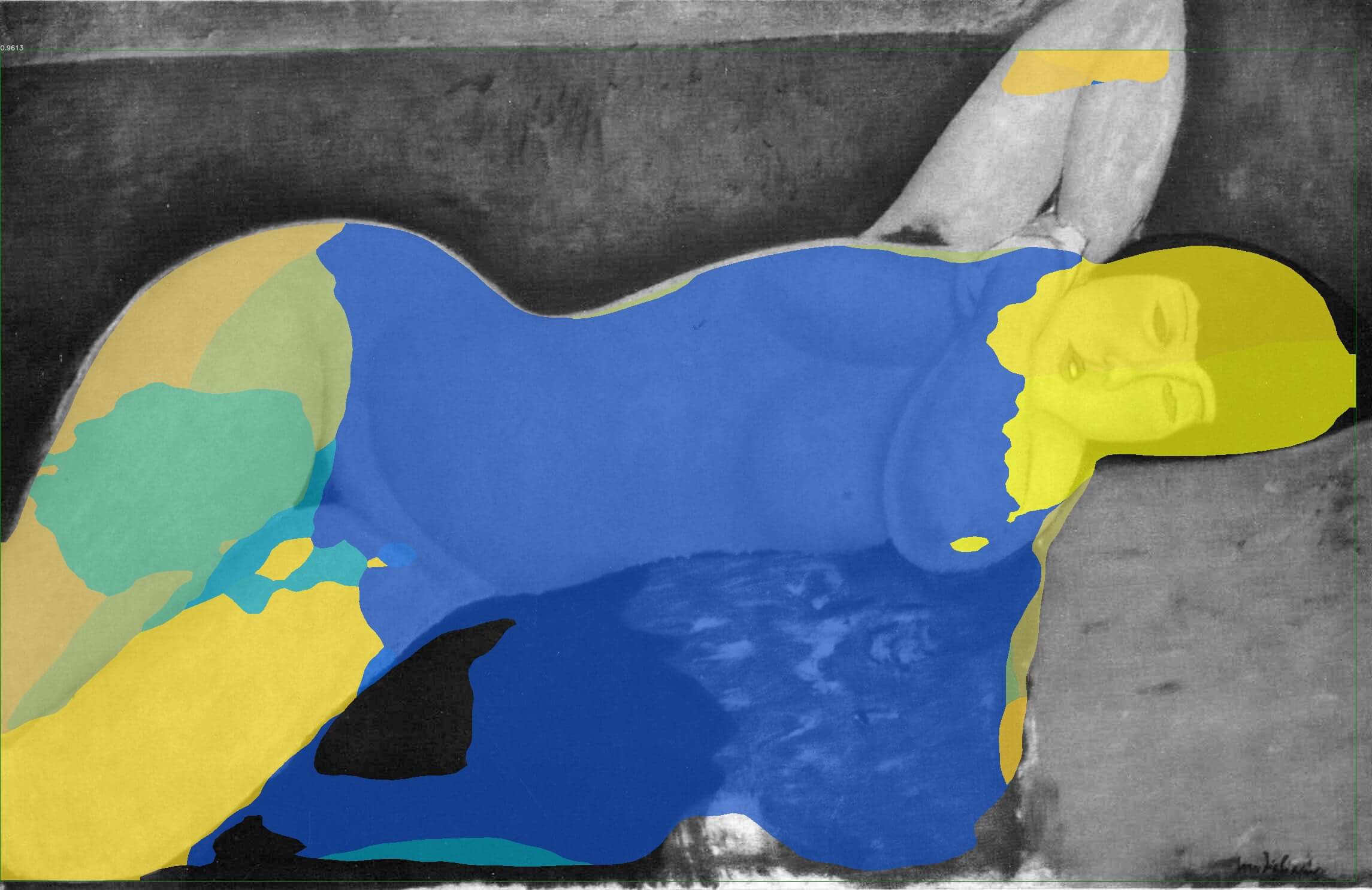}
  \includegraphics[height=3cm]{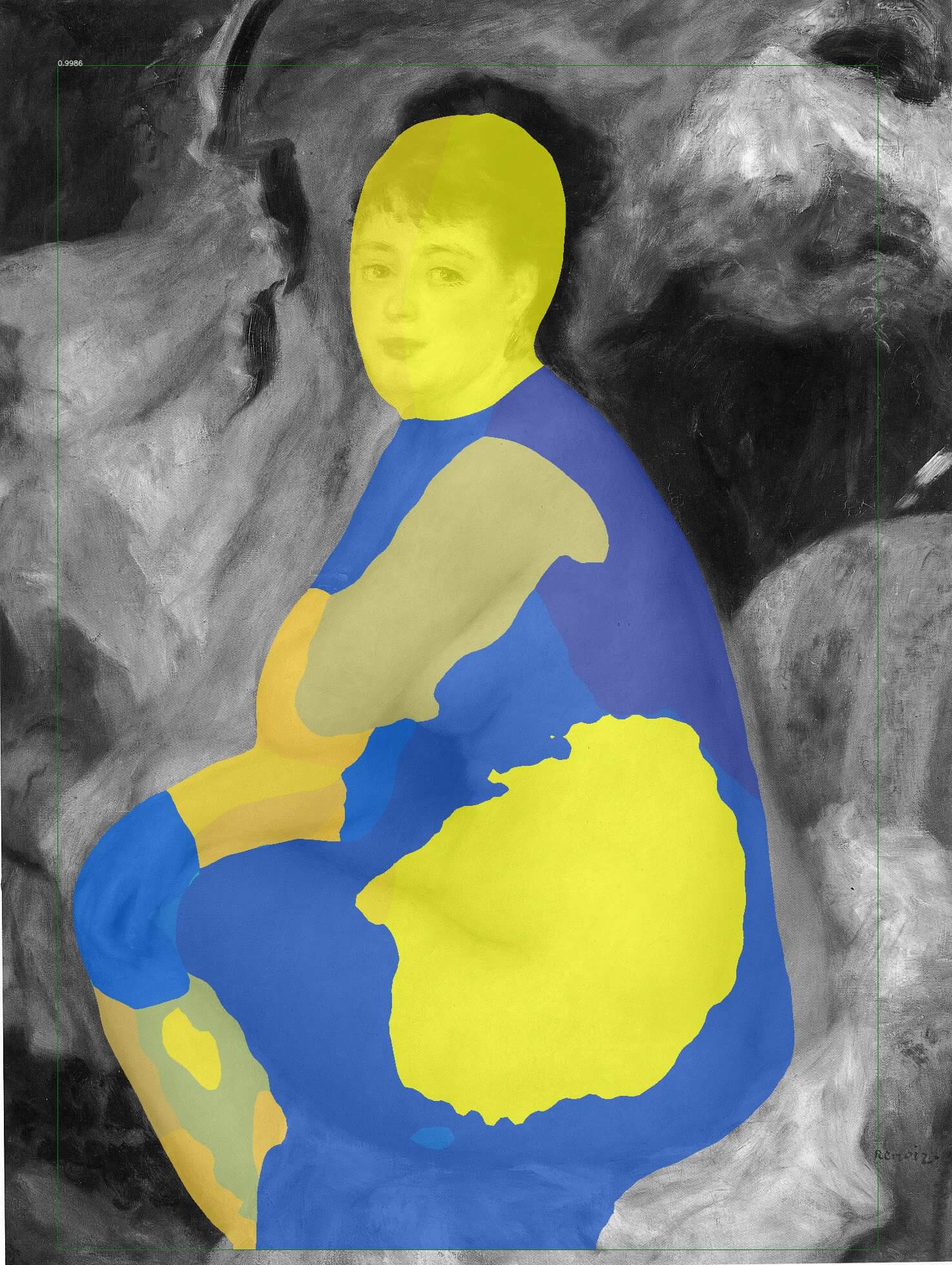}
  \includegraphics[height=3cm]{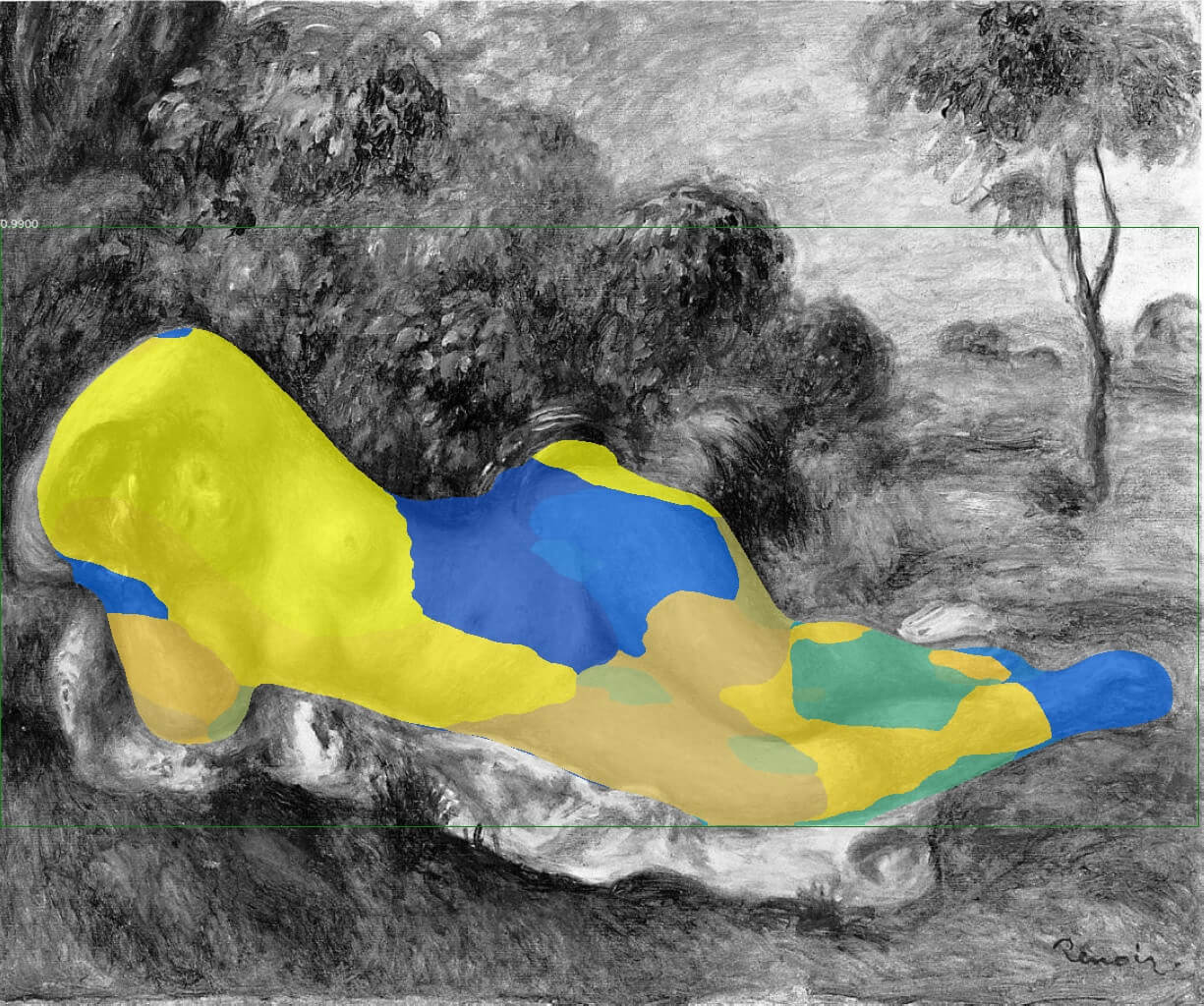}
  \includegraphics[height=3cm]{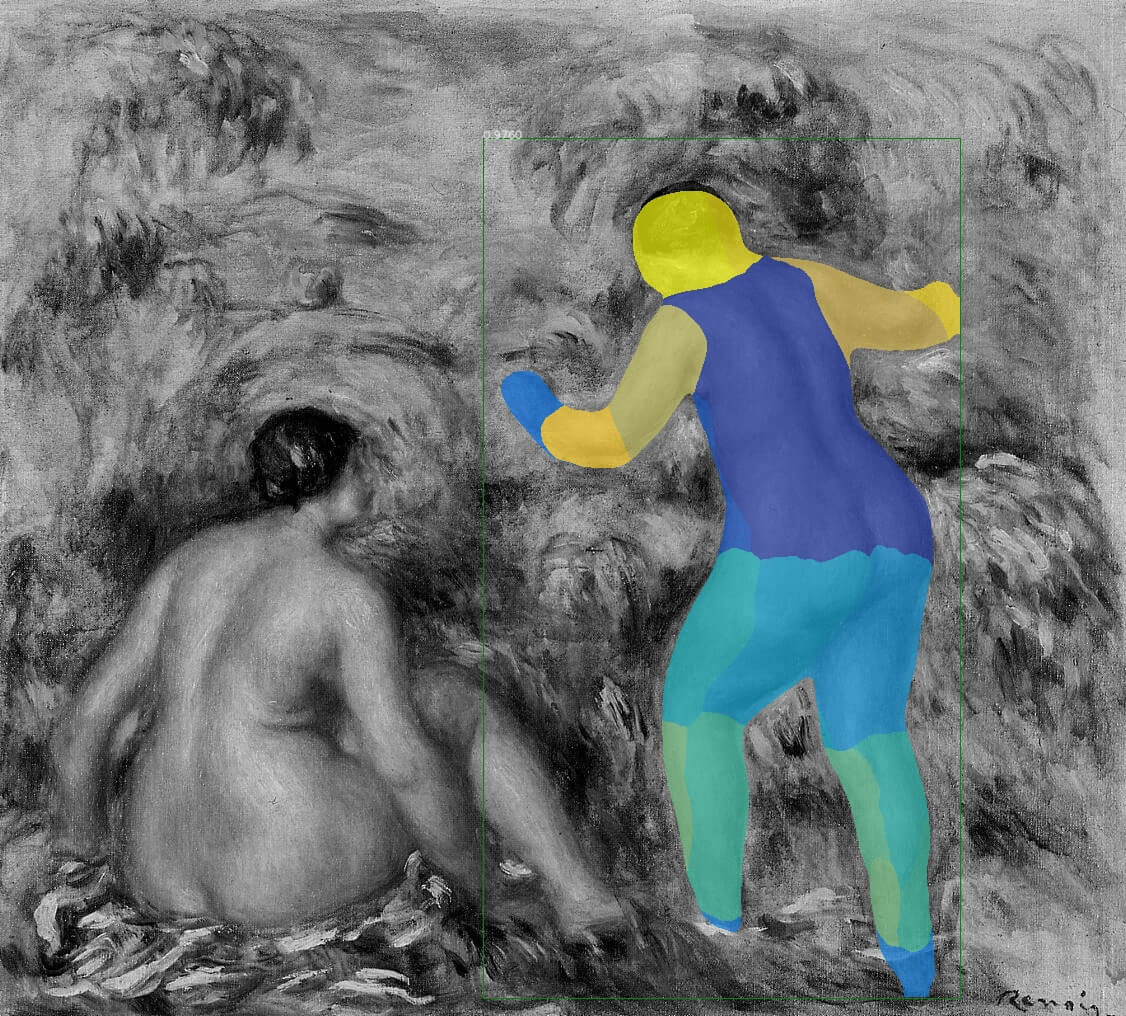}
  }
  
  \resizebox{.99\textwidth}{!}{       
  \includegraphics[height=3cm]{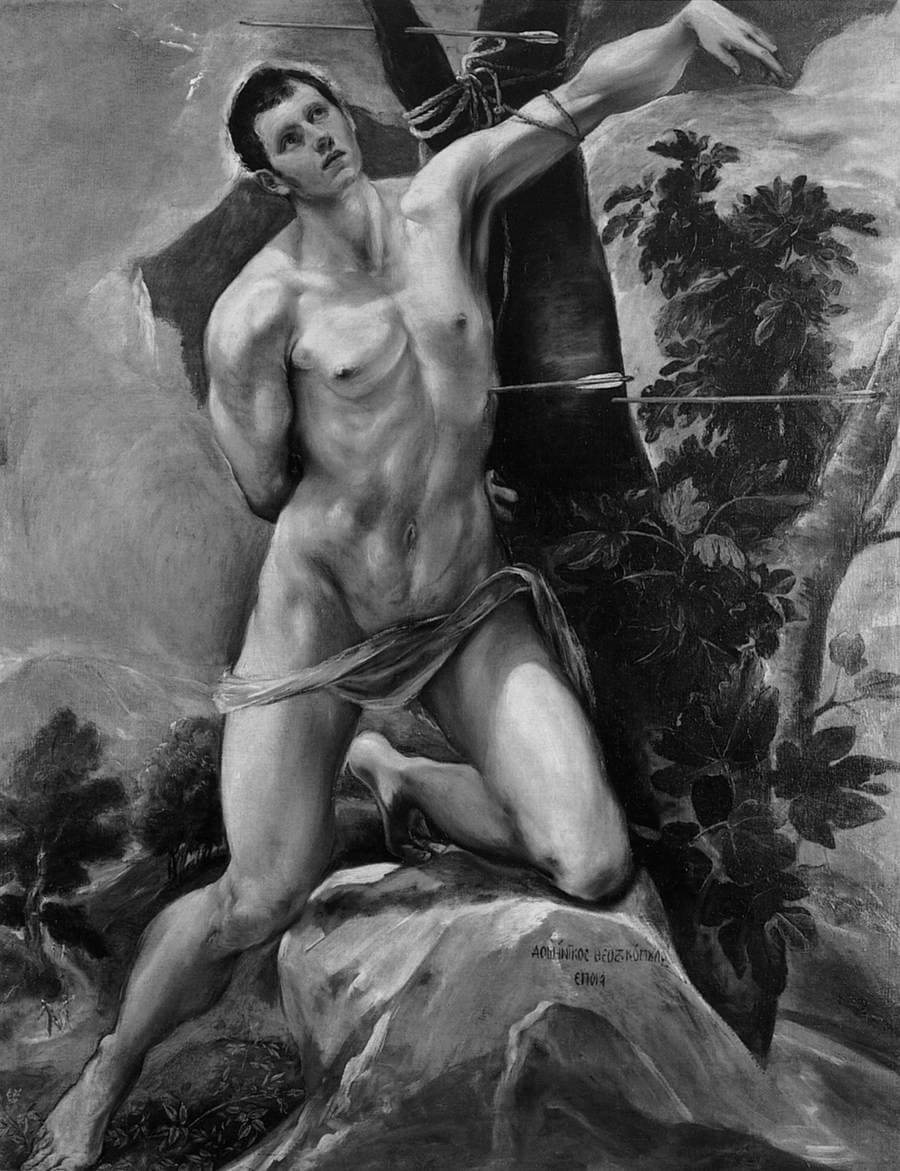}
  \includegraphics[height=3cm]{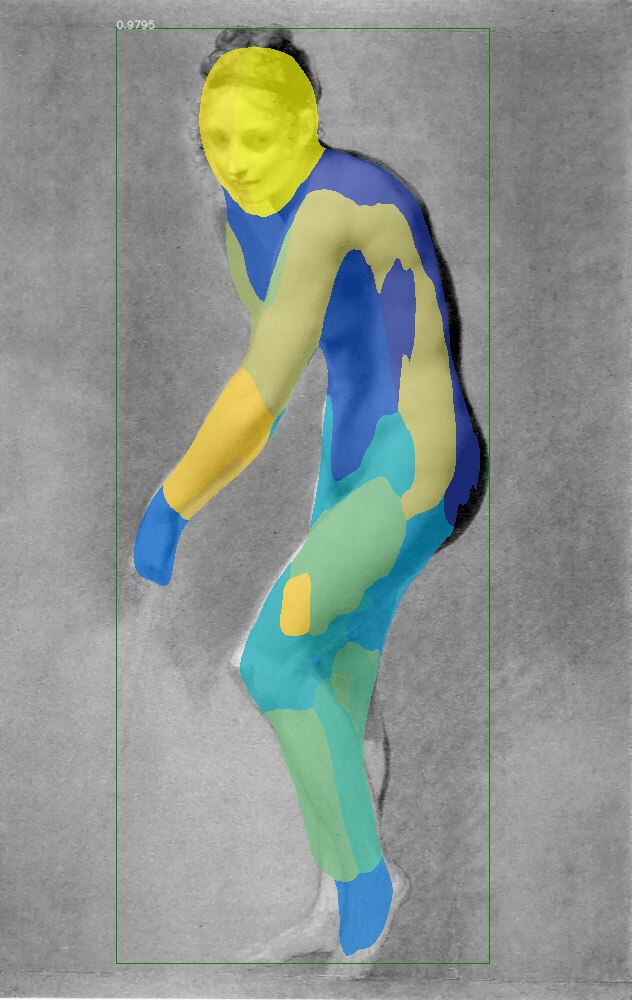}
  \includegraphics[height=3cm]{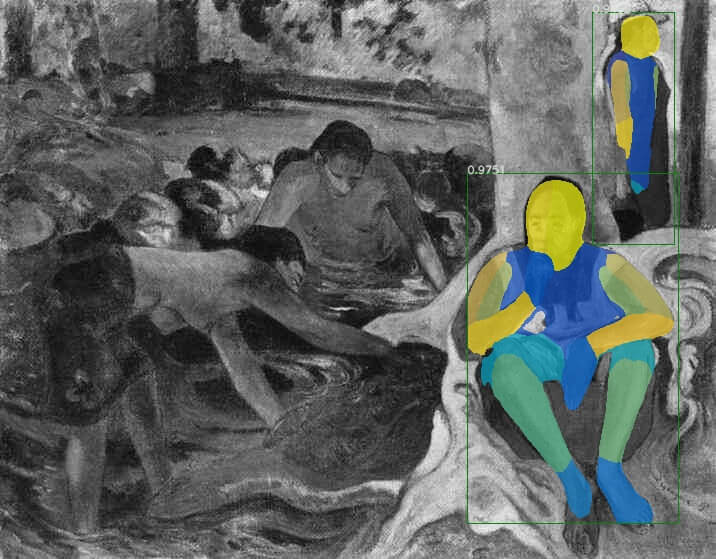}
  \includegraphics[height=3cm]{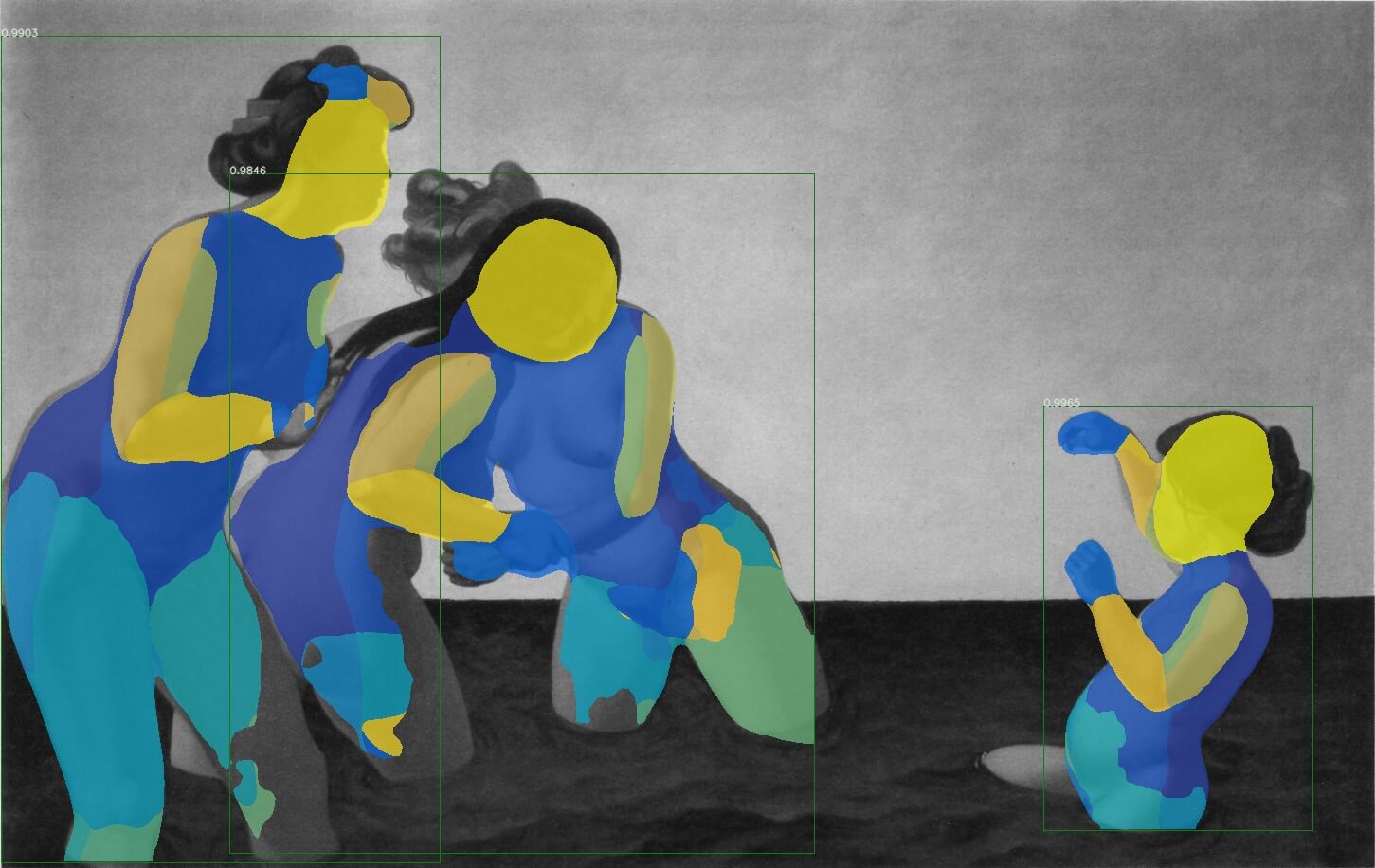}
  \includegraphics[height=3cm]{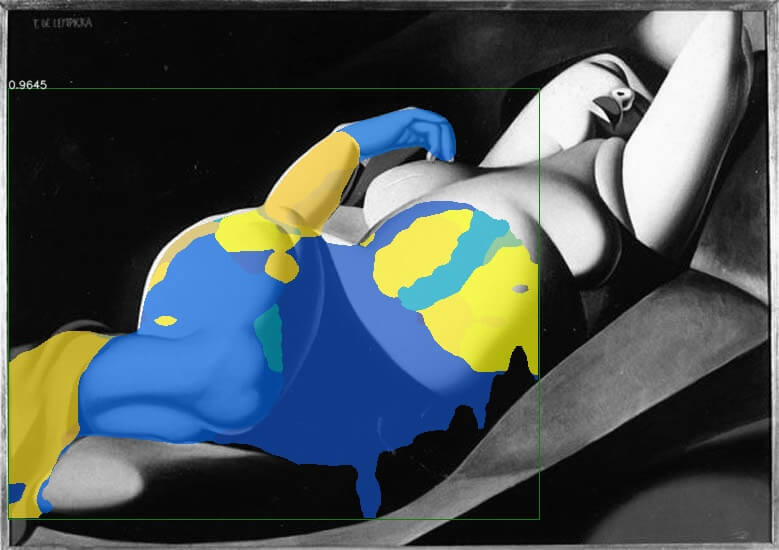}
  }
 \end{center}
  \caption{\label{fig:densepose}
           First row: The detection of body segments for Michelangelo (first three) and Paul Delvaux (last two).
           Middle row: The inference of segments for Modigliani (first two) and Renoir (last three).
            Last row, from left to right:
           The inference of segments for El Greco, Prud’hon, Gauguin, Vallotton and Lempicka.}
           
\end{figure}

The paintings of Michelangelo and Paul Delvaux have the highest accuracy. In their paintings, most of the poses are nude, hence without interference of clothes and in nude skin colours. For Michelangelo, we have one pose per painting to detect, and for Delvaux, the people are distanced from each other. All the poses from Michelangelo are sitting, and most poses from Delvaux are standing, which form the two most common pose groups, and all of their poses are depicted from a frontal view. Lastly, all of their poses follow natural body proportions. Figure~\ref{fig:densepose} shows successful and failed pose detections including these two artists. 

%The first one is the $100\%$ percent correctly inferred case for Michelangelo. The second one is also fully correct, but due to existence of the hanging robe, the legs cannot be inferred. The third one is a failure, and the potential reason might be that the torso is twisted and occluded by the right arm when viewed from aside, thus the torso is wrongly deduced, which further leads to the incorrect inference of lower limbs. The fourth one is the success case for Paul Delvaux, as most people painted by him are distanced between one another in order to create the lonesome ambience, the occlusion problem is largely avoided. The fifth one is the wrong case, as the lying pose constitutes one of the difficult poses. Even by rotating the lying pose $90$ degrees to a standing pose, DensePose still fails to infer correctly. The potential reason might be that the relative spatial relationships between body segments are different from those of a natural standing pose. In \Cref{sec:hierarchical-clustering}, it will further illustrate that the training dataset of DensePose has only very few lying natural poses as well. Thus, the lying poses are under-represented during training, which might lead to incorrect inference during test. Moreover, the sitting person in the background is missing in inference, as DensePose doesn't recognize this as a person, which reflects the problem of early commitment of top-down methods.

As shown in Table~\ref{tab:accuracy-keypoints}, the paintings of Amedeo Modigliani and Pierre-Auguste Renoir have the least accuracy, as most of their poses are very challenging. Lying poses of Modigliani, along with his use of overly slender and elongated body proportions are hard to detect. Similarly, El Greco's elongated bodies are not always easy to segment. If the bodies are painted by blobs of various colours (e.g. Renoir), or if the colour contrast is low (e.g. Gaugin) the inference suffers. The niche perspective poses are also a challenge for inference (e.g. Lempicka). Examples are given in Figure~\ref{fig:densepose}. 
%The first two are both failed cases for Modigliani: one is totally blank without any inference, the other is with wrong inference for torso, arm and right thigh. The third and fourth ones are failed cases for Renoir, and they might be due to his colored brushes, plus the twisted sitting pose and lying pose. The fifth one is a partially correct case, the sitting person is totally ignored, but for the standing pose, it is much easier to infer.

%For the other artists, the examples are shown in \autoref{fig:densepose-other-cases}. For El Greco, the elongated body cannot be inferred for its segments. For Prud’hon, the standing pose viewed from aside is hard to infer, and especially, the torso is wrongly masked. For Gauguin, color contrast might impact inference. The man sitting closely behind and the woman hunching her back are totally missed. For Felix Vallotton, the intertwined arms are incorrectly inferred. For Lempicka, the niche perspective poses a challenge for inference. Moreover, the exaggerated body segments are themselves hard to infer, in which upper limbs have comparatively higher accuracy than that of torso and lower limbs.

The root causes of these failures may include the fact that the COCO dataset used in the training of these models has a lower density of natural poses, such as lying, sitting with back turned to observers, crowds of overlapping people, and such. Instead, many training samples depict single persons, doing sports or with only the upper torso captured. Furthermore, the training data collects mostly common people with common body proportions and height, which are neither too slender, nor too plump. Yet artistic depictions can go to extremes.

\subsection{Average Contours of Body Segments}
\begin{figure}[thb!]
  \centering
   \resizebox{\textwidth}{!}{
  \includegraphics[height=2cm]{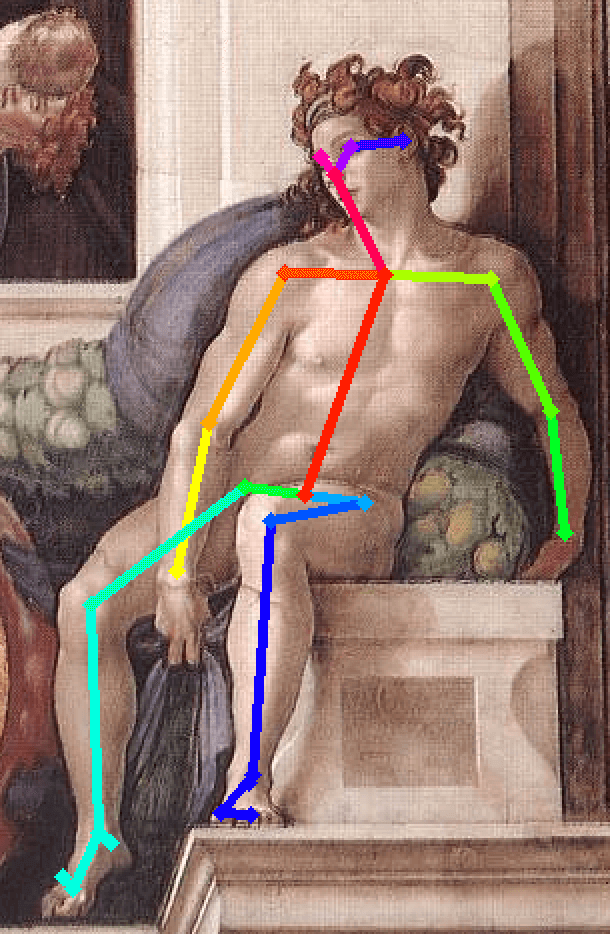}
  \includegraphics[height=2cm]{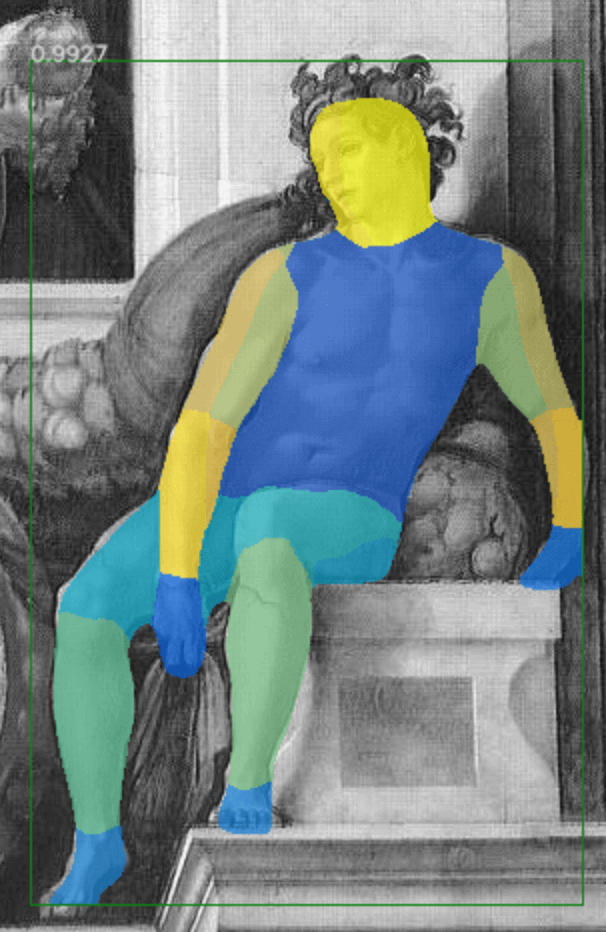}
  \includegraphics[height=2cm]{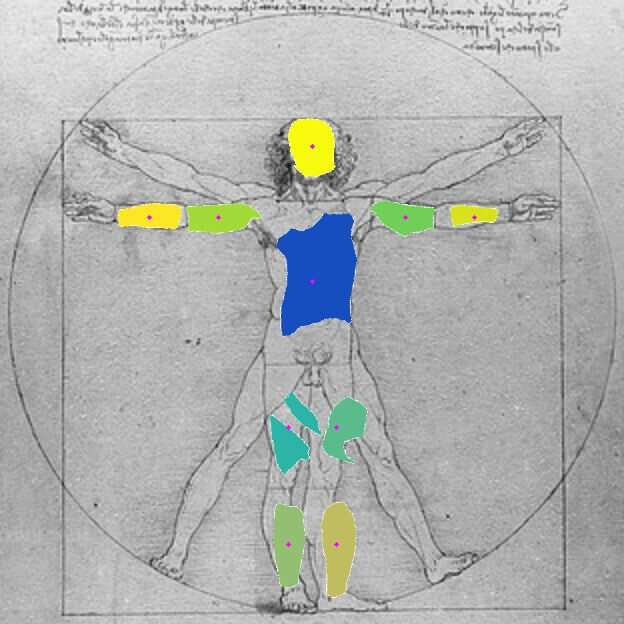}
  \includegraphics[height=2cm]{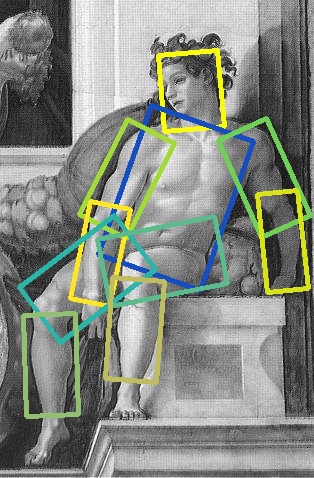}
  }
  \centering
  \resizebox{\textwidth}{!}{
  \includegraphics[height=2cm]{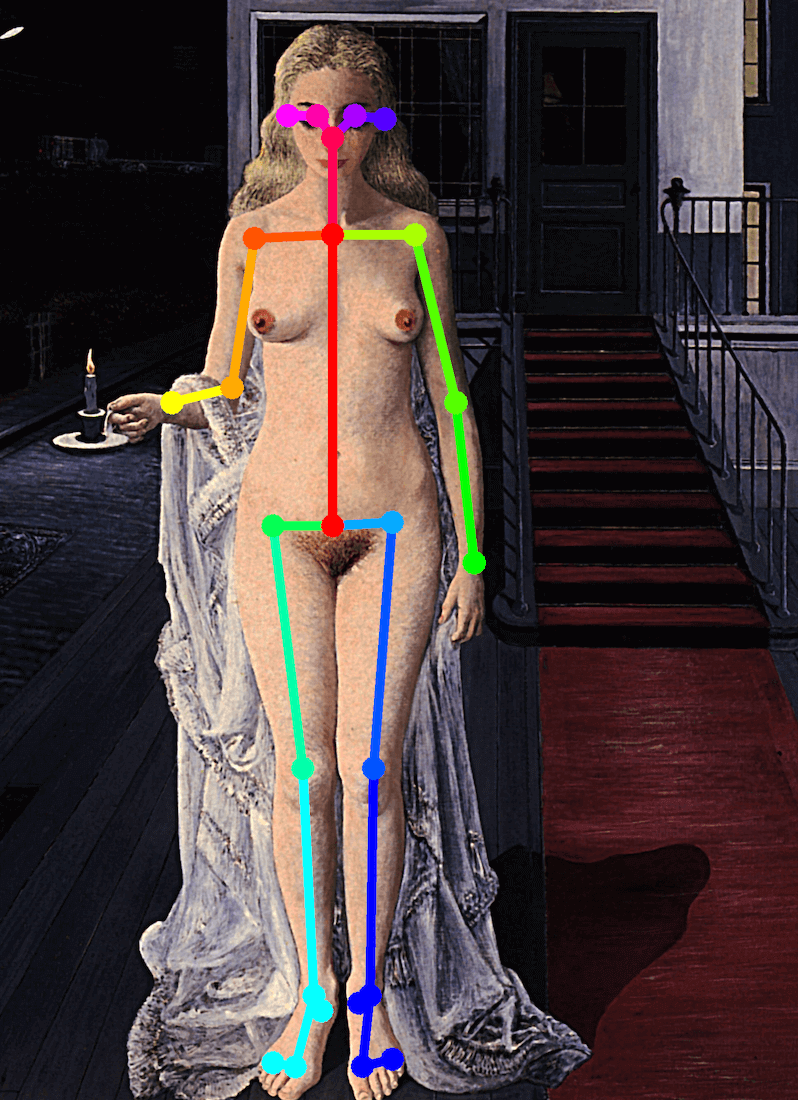}
  \includegraphics[height=2cm]{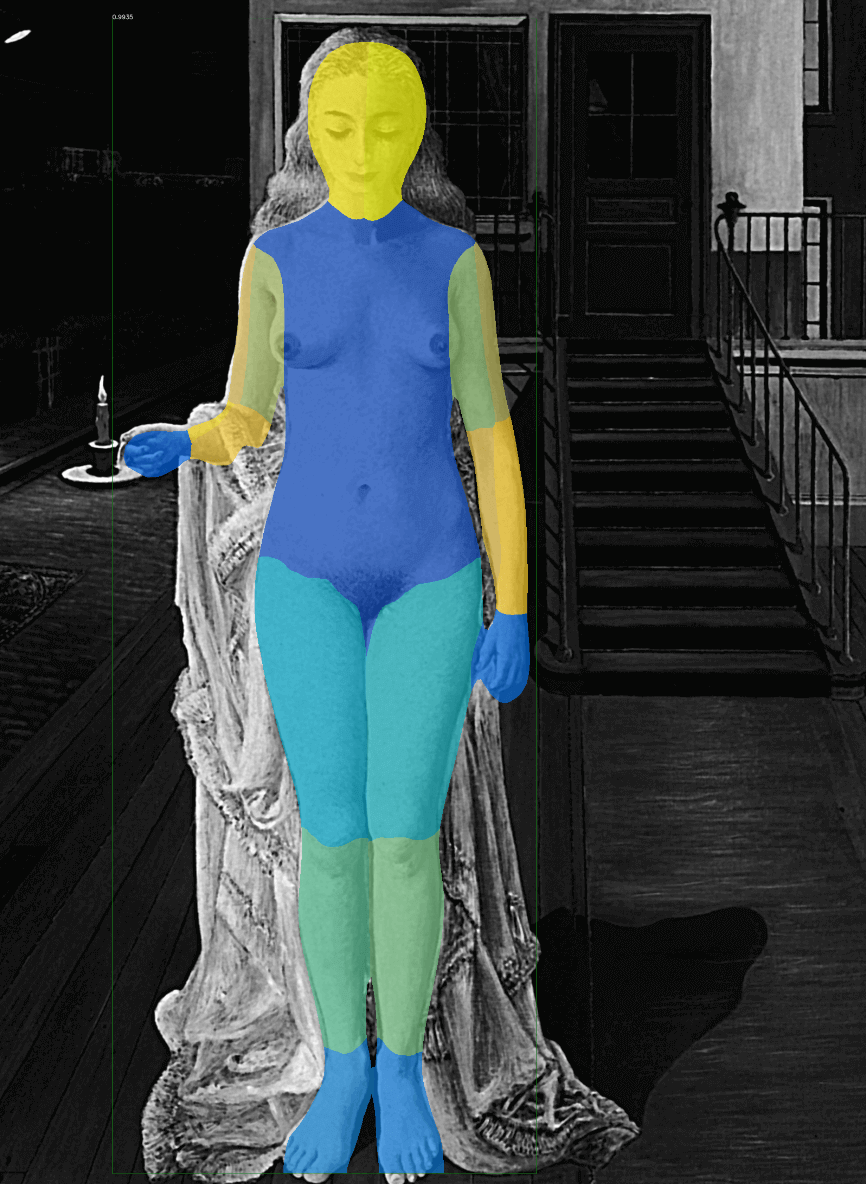}
  \includegraphics[height=2cm]{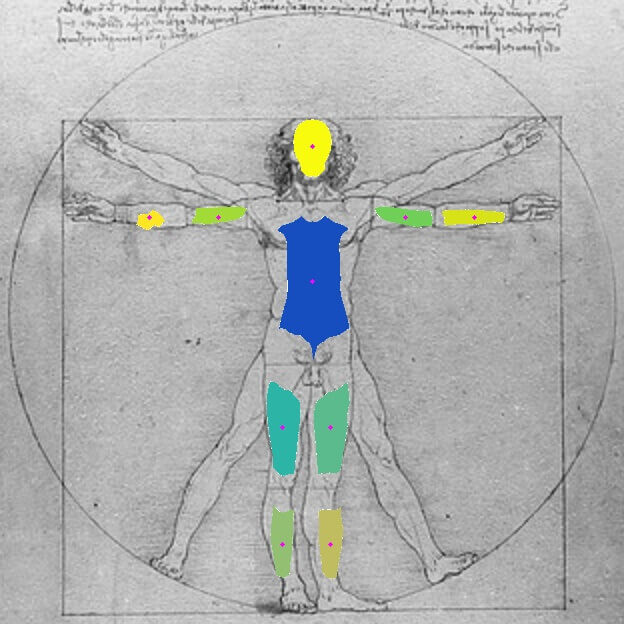}
  \includegraphics[height=2cm]{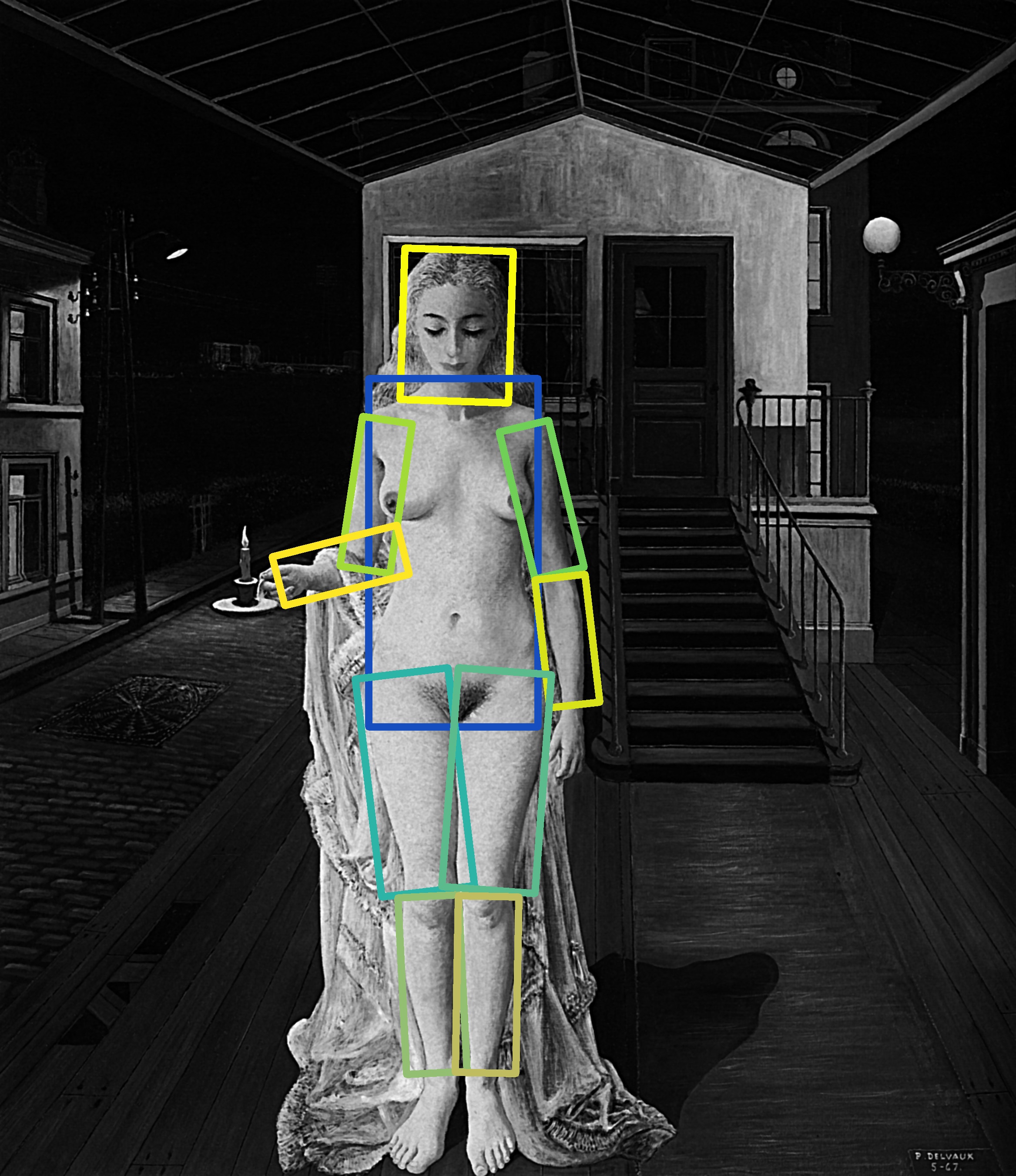}
  }
  
  \centering
   \resizebox{\textwidth}{!}{
  \includegraphics[height=2cm]{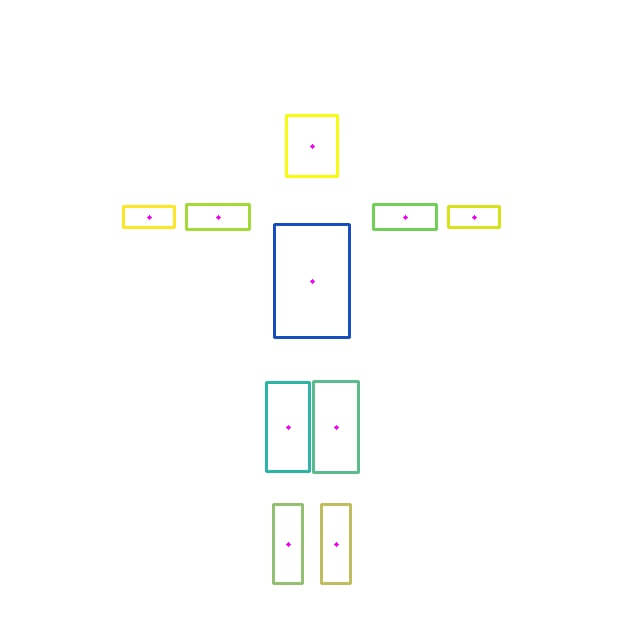}
  \includegraphics[height=2cm]{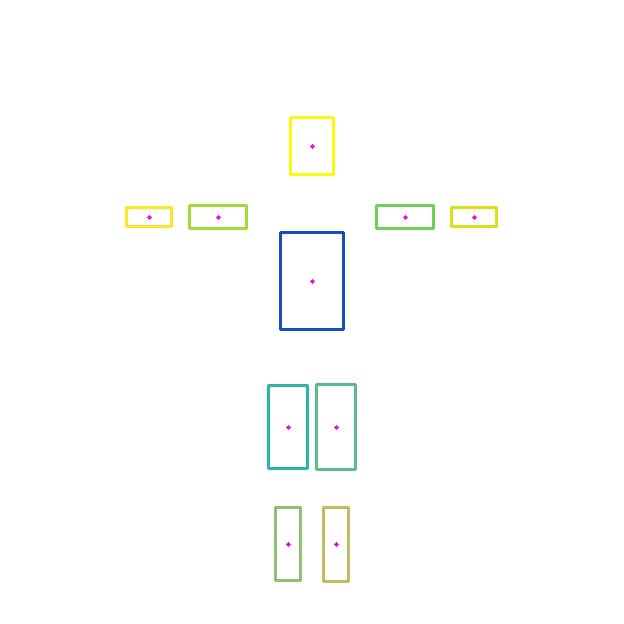}
  \includegraphics[height=2cm]{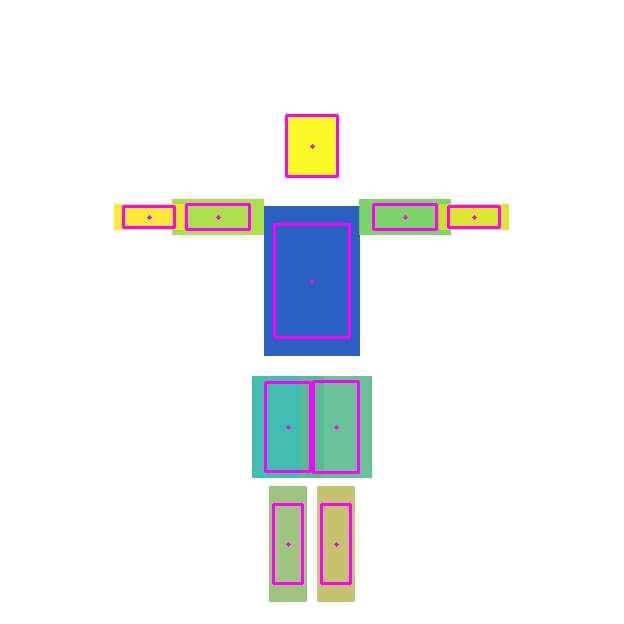}
  \includegraphics[height=2cm]{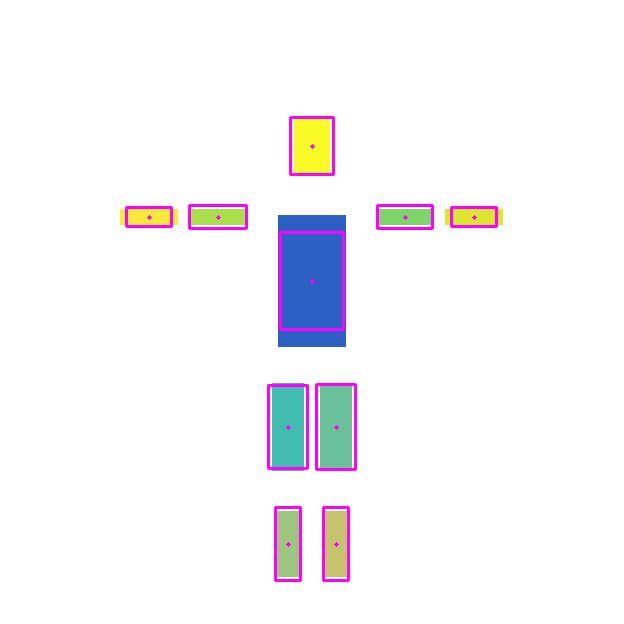}
  }
  
 \caption{\label{fig:vitruvian man}
           Michelangelo's man (first row) and Paul Delvaux's woman (middle row). For both rows, left to right: Keypoints by OpenPose, segments by DensePose, normalised DensePose on Vitruvian Man and on the original image. Last row, left to right: 1) Average contour of the COCO men. 2) Average contour of the COCO women. 3) Michelangelo pose imposed on the contour of the COCO men. 4) The Delvaux pose imposed on the contour of the COCO women.}
           
\end{figure}

We use Michelangelo and Paul Delvaux as examples to demonstrate the normalisation process and the results (see Figure~\ref{fig:vitruvian man}). When we compare these two normalised poses, we see that the body drawn by Michelangelo is more inflated and fills to the brim the under the contours of the Vitruvian Man, whereas Paul Delvaux's woman is slender. Compared with Paul Delvaux's standing pose, Michelangelo's sitting pose exposes three major issues: (1) The right thigh is occluded by the right arm, which leads to the partial segment with a gap in the normalised pose. (2) The thighs are not fully stretched out due to the sitting pose and the viewing perspective. Thus, the corresponding normalised thighs are both shorter. (3) The lower left arm is retracted a little behind the torso, which is further away from the observer. Thus, the corresponding normalised lower left arm is shorter and thinner, compared to the lower right arm.

Figure \ref{fig:vitruvian man} shows the average contours of the COCO men and women respectively, calculated with the same approach, using 144 men and 150 women from the COCO Person dataset. Superimposing the same pose from Michelangelo and Paul Delvaux on the natural men and women contours shows that Michelangelo tends to exaggerate the muscles of men, as every segment is inflated outside the brim. On the contrary, Paul Delvaux tends to draw the women more slender than their counterparts in the natural poses, as the torso and the limbs shrink a bit width-wise within the borders (see Figure~\ref{fig:vitruvian man}).

In summary, the mean contours can give us an intuitive view of artistic and natural poses,
focusing on height and width of body segments. For art historical analysis, the segment contours of each artist can be visualised as to how each artist tends to draw human bodies. The contours can be compared with each other to further explore whether there exist significant differences between the drawing styles for each artist, and with that of the COCO men and women contours to analyse whether the artistic contour conforms to or deviates from the natural contours. %For computational purposes, the average contours can be used as the bounding box for each segment in order to enhance the inference accuracy of DensePose. If some segments are occluded and hence only partially inferred in one pose, they can be compensated from other poses. However, as demonstrated in the example of Michelangelo, this approach may render unreliable results if the chosen artist has mainly drawn sitting or other niche poses. 
%To albert: cok mu tekrar var?

\subsection{Visualising joint distributions}
\label{subsection:ellipsoids}
\begin{figure}

  \centering
  \resizebox{\textwidth}{!}{
  \includegraphics[height=4cm]{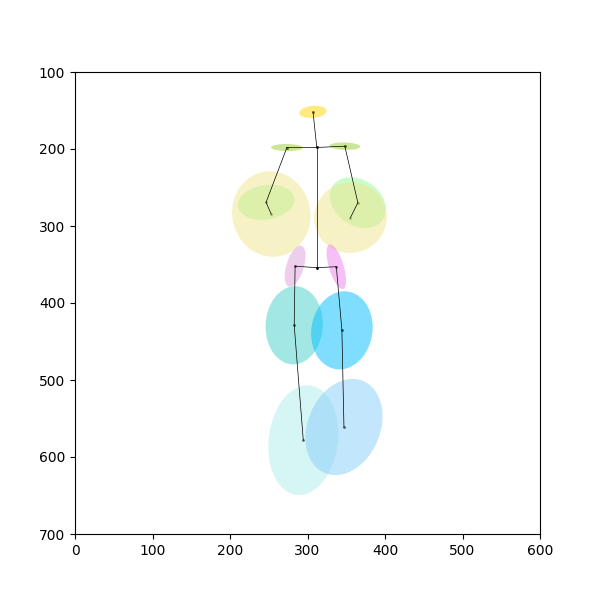}
    \includegraphics[height=4cm]{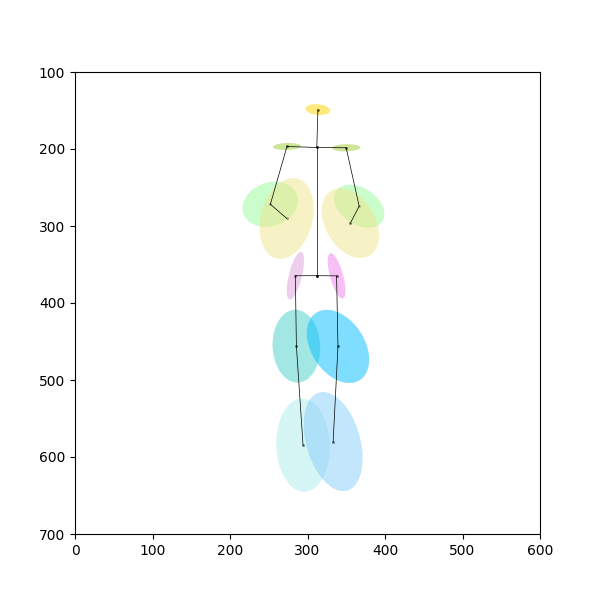}
    \includegraphics[height=4cm]{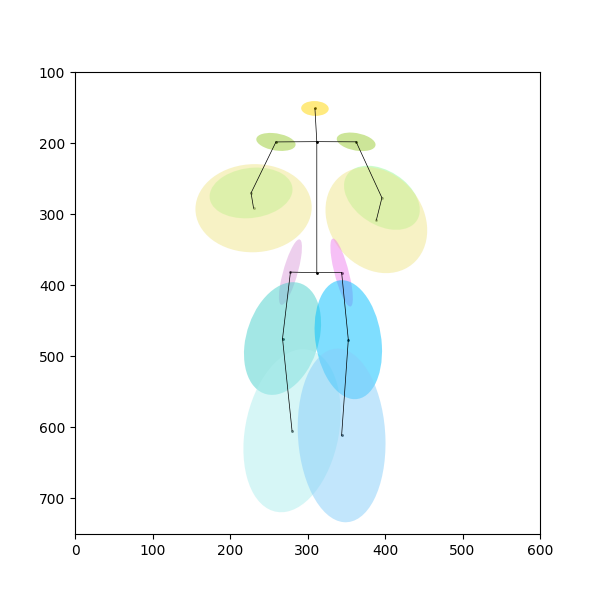}
    \includegraphics[height=4cm]{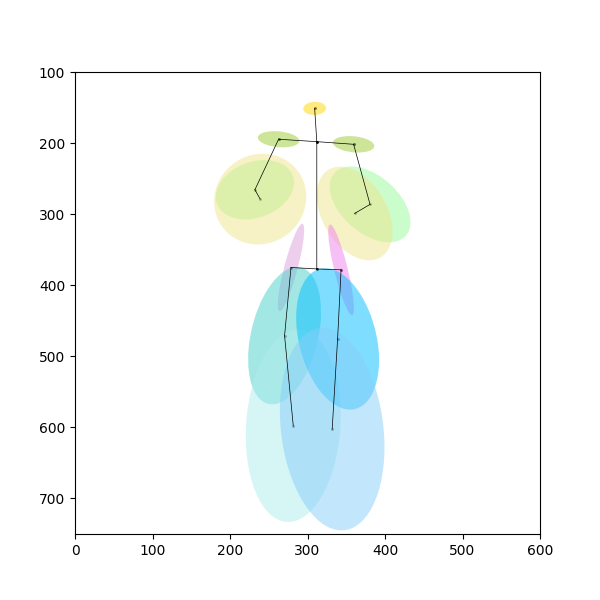}
    }
  \caption{
            The mean and $0.5$ standard deviation distribution for artistic (AP) and natural poses (COCO). Joints used in the normalisation have smaller variances. From left to right: Males in AP; Females in AP; Males in COCO; Females in COCO.} 
            \label{fig:ellipse}
\end{figure}

For the AP dataset, after the first step of the validity check, we have $211$ poses. Out of these, 67 poses are male, and 144 female (Figure~\ref{fig:ellipse}). Different from the distributions of facial landmarks~\cite{yaniv2019face}, the keypoints are prone to high variance of articulation. For the upper limbs, the wrists occupy a larger ellipsoid than the elbows, as do the elbows compared to the shoulders. Similarly, for the lower limbs, the areas of the ellipsoids are in descending order: ankle $>$ knee $>$ hip. In general, the outer joints swing over a larger circle than the inner joints. We also observe that the joint distributions are similar for both male and female poses, as they are both with arms and legs hanging alongside the torso.

As a reference, we also analyse the gestures of daily activities captured in photographs. After filtering through the COCO people dataset and a validity check, we use $653$ male and $222$ female poses for the mean and standard deviation analysis. Compared with artistic poses, natural poses tend to be more varied with wider movement of joints. The natural poses are formed by a wide variety of daily activities and sports, whereas the artistic poses are staged and hence restrained to certain poses (see Figure~\ref{fig:ellipse}).

\subsection{Hierarchical clustering}
\label{subsection:hierarchical}
\begin{figure}
  \centering
  \includegraphics[width=0.9\linewidth]{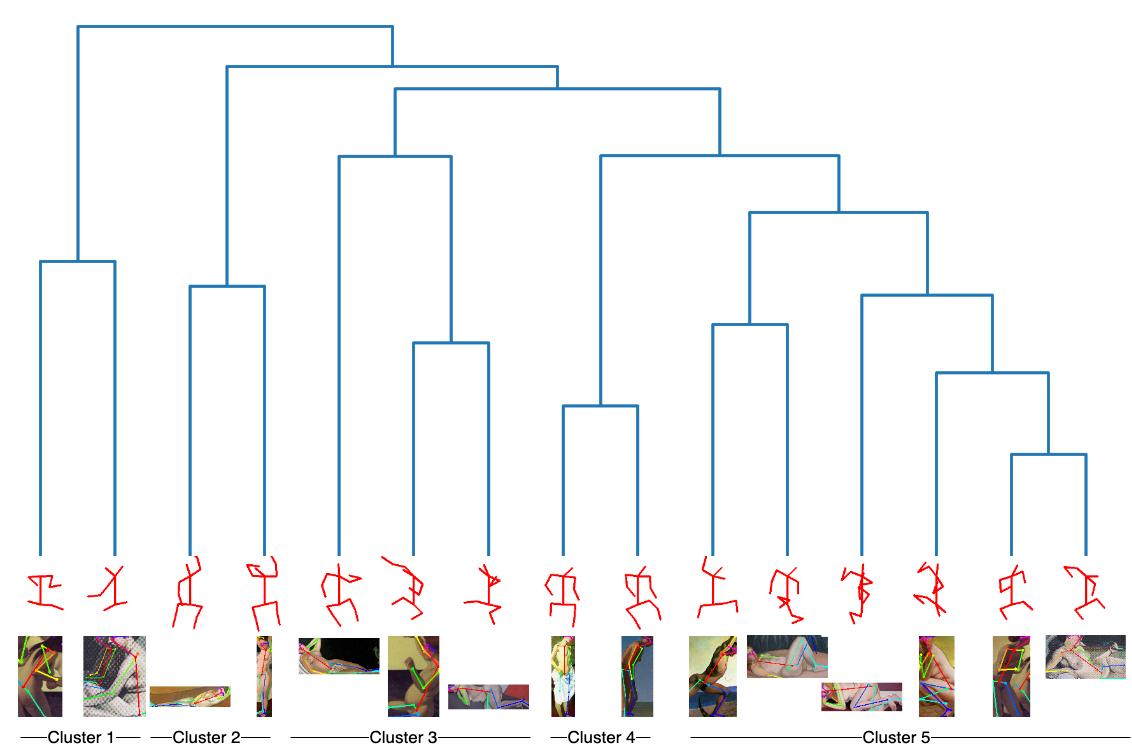}
  \caption{\label{fig:clusters-felix-vallotton}
           Dendrogram for all $15$ poses drawn by Felix Vallotton from $15$ paintings.}
\end{figure}

We use the poses from the paintings of Felix Vallotton as an example to demonstrate the hierarchical clustering results. Among the $15$ full-body depictions of Felix Vallotton figures, one can find variations such as lying, sitting, and standing poses. The hierarchical clustering, as shown in Figure~\ref{fig:clusters-felix-vallotton}, cannot distinguish whether a person lies down, sits or stands. But the clusters can differentiate between the stretched and compressed limbs. When the number of clusters is set to $5$, cluster 2 and 3 contain the poses with arms thrown up; cluster 4 contains the standing poses with arms relaxed alongside the body; cluster 5 contains the twisted legs and arms.

\begin{figure}[htb]
  \centering
  \includegraphics[width=0.8\linewidth]{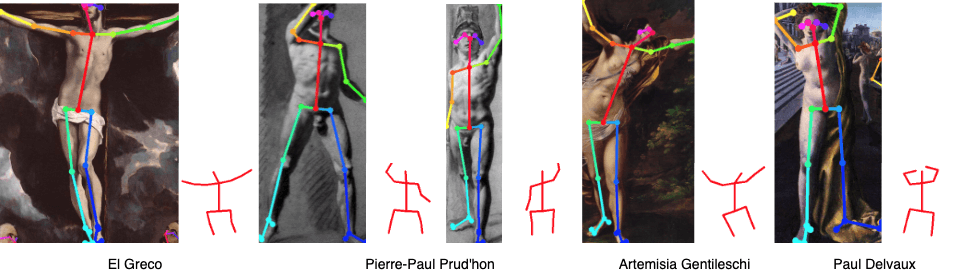}
    \includegraphics[width=1.0\linewidth]{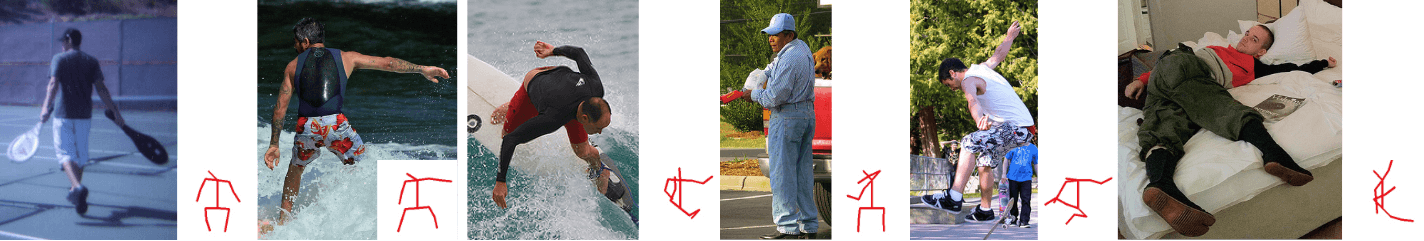}
  \includegraphics[width=1.0\linewidth]{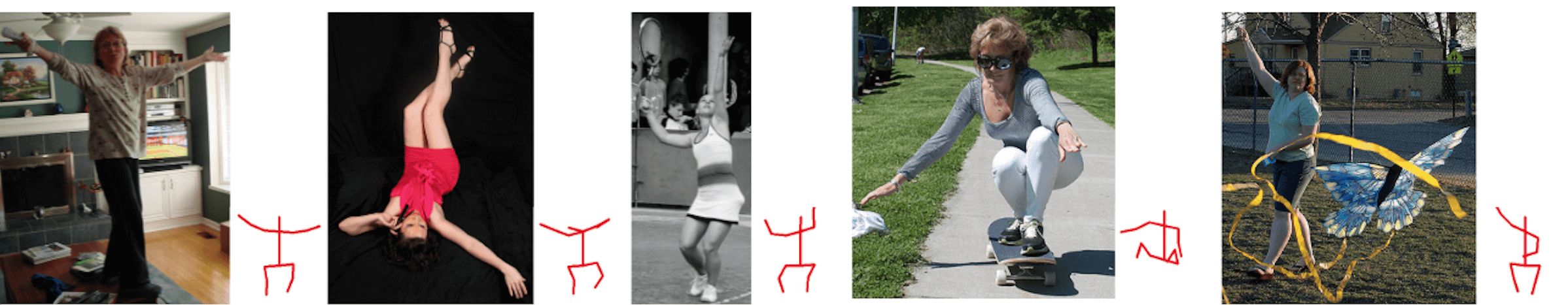}
    \caption{\label{fig:clusters-arms-upward}
    First row: The standing poses with arms thrown upward into the air, as an example niche pose from the AP dataset. Middle row: The niche natural poses in the COCO dataset for only men. Last row: The niche natural poses in the COCO dataset for only women.
}
\end{figure}

When the number of clusters is set to $10$ in order to generate the hierarchical clustering for all the artists, the outcome shows that there are no clear boundaries between artists. Out of these $10$ clusters, there are $2$ outstanding clusters with only a small fraction of poses. %These poses are shown in Figure \ref{fig:clusters-arms-upward}, where the left image shows the original pose, and the right image shows the normalized poses with all limbs equal to $30$ pixels. %The normalized hip width is equal to the length of two limbs, which seems a bit wide, but it can better illustrate the poses with crossed legs. 
These are all standing poses with the arms thrown upward. The pose of opening arms appears for example in the religious paintings of El Greco, but similar poses are depicted by Artemisia Gentileschi, Pierre-Paul Prud'hon and is later used by Paul Delvaux as well (see Figure~\ref{fig:clusters-arms-upward}). %The lying nude women as in those of Amedeo Modigliani depicts the same pose. 

For comparison, we carried out hierarchical clustering in the COCO dataset to find niche poses for men and women. The rare poses for men are usually with more stretched arms, i.e., swinging back and forth, or the ones with more twisted legs. Moreover, the lying poses barely appear for men. The rare poses for women are the ones with twisted or overly stretched legs, or those with arms thrown upward (Figure
~\ref{fig:clusters-arms-upward}).

In summary, the hierarchical clustering of the pose vectors can tell the difference of poses in: (1) the orientation, (2) whether the limbs are stretched or compressed, and (3) the completeness of the joints. Via this visualisation method it is also possible to track the transformation and re-use of similar poses by different artists over time. When we compare the artistic and natural poses we see that the poses with arms stretched upward are quite rare for natural poses as they appear only in women's hugging or lying poses, or in sports. It might be assumed that if trained with more niche poses and niche perspectives from natural poses, OpenPose and DensePose might also perform better at inference of artistic poses.

\section{Conclusions}
\label{section:conclusions}

In this paper, we evaluated automated human body analysis for paintings in Western artworks. Specifically, we tested two state of the art pose estimation models, namely DensePose and OpenPose, on a curated and manually annotated artistic dataset of 10 Western Artists from various genres and periods, and reported their performances. OpenPose in general performed better than DensePose, especially for paintings with more than one figure where figures are interacting with each other. There is room for improvement in pose segmentation, and fine-tuning on annotated painting datasets may help.
%Unlike previous approaches in literature, to annotate this dataset we focused on a semi-automatic annotation method whereas we first applied the out of the shelf methods without any pre-training. We annotated only the failed cases/keypoints where the inference of PCK suffers.  

We used a simple, rule-based approach to improve the performance of DensePose by combining it with the keypoints of OpenPose, and proposed a way of visualising a painter's preferred body shapes using a normalisation method using Leonardo's Vitruvian Man. This way, we generate average contours for all poses drawn by a given artist, which can also be used to summarise differences in body shape preferences of artists. For example, a body that looks fat with respect to an average body can be classified as skinny with respect to the artist's specific style. The same holds true for other shape parameters. Our proposed approach improves over purely skeleton-based analysis of human poses.

%-vitrivian mani yaptik bu yontemle her sanatcinin kullandigi ortalama body shapei de gorebiliyor ve birbiri ile kiyasliyabiliyoruz, ve bu yontemi natural poselara uygulayarak ve karsilastirma yaparak sanatsal pozlar ve gundelik pozlar arasindaki oran ve body shape (siskolik incelik uzunluk) gorebiliyoruz

%pose analizini 3d shape uzerinden yaptik, daha once sadece keypointlar arastirilmisti, oysa ki body shape keypoint analizinde kayboluyor

Finally, we have analysed the distribution of joints in natural photographs and paintings, and shown how our approach can automatically find niche poses preferred by artist. The niche poses we found in the AP dataset via the agglomerative hierarchical clustering are also documented in Impett and Susstrunk's study of Aby Warburg’s Bilderatlas~\cite{impett2016pose}, which shows that our automatic approach is useful. Our findings also point out to what kind of poses are needed for the training of pose detectors for paintings.

\bibliographystyle{splncs04}
\bibliography{zhao22visart}

\begin{thebibliography}{10}
\providecommand{\url}[1]{\texttt{#1}}
\providecommand{\urlprefix}{URL }
\providecommand{\doi}[1]{https://doi.org/#1}

\bibitem{aviezer2012body}
Aviezer, H., Trope, Y., Todorov, A.: Body cues, not facial expressions,
  discriminate between intense positive and negative emotions. Science
  \textbf{338}(6111),  1225--1229 (2012)

\bibitem{bai2021explain}
Bai, Z., Nakashima, Y., Garcia, N.: Explain me the painting: Multi-topic
  knowledgeable art description generation. In: Proceedings of the IEEE/CVF
  International Conference on Computer Vision. pp. 5422--5432 (2021)

\bibitem{cao2019openpose}
Cao, Z., Hidalgo, G., Simon, T., Wei, S.E., Sheikh, Y.: Openpose: realtime
  multi-person 2d pose estimation using part affinity fields. IEEE transactions
  on pattern analysis and machine intelligence  \textbf{43}(1),  172--186
  (2019)

\bibitem{castellano2021visual}
Castellano, G., Lella, E., Vessio, G.: Visual link retrieval and knowledge
  discovery in painting datasets. Multimedia Tools and Applications
  \textbf{80}(5),  6599--6616 (2021)

\bibitem{castellano2021deep}
Castellano, G., Vessio, G.: Deep learning approaches to pattern extraction and
  recognition in paintings and drawings: An overview. Neural Computing and
  Applications  \textbf{33}(19),  12263--12282 (2021)

\bibitem{cetinic2021iconographic}
Cetinic, E.: Iconographic image captioning for artworks. In: International
  Conference on Pattern Recognition. pp. 502--516. Springer (2021)

\bibitem{Guler2018DensePose}
G\"uler, R.A., Neverova, N., Kokkinos, I.: {DensePose}: Dense human pose
  estimation in the wild. In: The IEEE Conference on Computer Vision and
  Pattern Recognition (CVPR) (2018)

\bibitem{huang2017arbitrary}
Huang, X., Belongie, S.: Arbitrary style transfer in real-time with adaptive
  instance normalization. In: Proceedings of the IEEE International Conference
  on Computer Vision. pp. 1501--1510 (2017)

\bibitem{impett2016pose}
Impett, L., S{\"u}sstrunk, S.: Pose and pathosformel in {Aby Warburg’s
  Bilderatlas}. In: European Conference on Computer Vision. pp. 888--902.
  Springer (2016)

\bibitem{jenicek2019linking}
Jenicek, T., Chum, O.: Linking art through human poses. In: 2019 International
  Conference on Document Analysis and Recognition (ICDAR). pp. 1338--1345. IEEE
  (2019)

\bibitem{king2009dlib}
King, D.E.: Dlib-ml: A machine learning toolkit. The Journal of Machine
  Learning Research  \textbf{10},  1755--1758 (2009)

\bibitem{langfeld2018canon}
Langfeld, G.: The canon in art history: concepts and approaches. Journal of Art
  Historiography (19),  152--180 (2018)

\bibitem{lin2014microsoft}
Lin, T.Y., Maire, M., Belongie, S., Hays, J., Perona, P., Ramanan, D.,
  Doll{\'a}r, P., Zitnick, C.L.: Microsoft coco: Common objects in context. In:
  European conference on computer vision. pp. 740--755. Springer (2014)

\bibitem{madhu2020understanding}
Madhu, P., Marquart, T., Kosti, R., Bell, P., Maier, A., Christlein, V.:
  Understanding compositional structures in art historical images using pose
  and gaze priors. In: European Conference on Computer Vision. pp. 109--125.
  Springer (2020)

\bibitem{madhu2020enhancing}
Madhu, P., Villar-Corrales, A., Kosti, R., Bendschus, T., Reinhardt, C., Bell,
  P., Maier, A., Christlein, V.: Enhancing human pose estimation in ancient
  vase paintings via perceptually-grounded style transfer learning. arXiv
  preprint arXiv:2012.05616  (2020)

\bibitem{magazu2019vitruvian}
Magaz{\`u}, S., Coletta, N., Migliardo, F.: The {Vitruvian Man of Leonardo da
  Vinci} as a representation of an operational approach to knowledge.
  Foundations of Science  \textbf{24}(4),  751--773 (2019)

\bibitem{murtinho2015leonardo}
Murtinho, V.: Leonardo’s {Vitruvian Man} drawing: A new interpretation
  looking at {Leonardo’s} geometric constructions. Nexus Network Journal
  \textbf{17}(2),  507--524 (2015)

\bibitem{noroozi2018survey}
Noroozi, F., Kaminska, D., Corneanu, C., Sapinski, T., Escalera, S.,
  Anbarjafari, G.: Survey on emotional body gesture recognition. IEEE
  transactions on affective computing  (2018)

\bibitem{ren2015faster}
Ren, S., He, K., Girshick, R., Sun, J.: Faster r-cnn: Towards real-time object
  detection with region proposal networks. arXiv preprint arXiv:1506.01497
  (2015)

\bibitem{sagonas2013300}
Sagonas, C., Tzimiropoulos, G., Zafeiriou, S., Pantic, M.: 300 faces
  in-the-wild challenge: The first facial landmark localization challenge. In:
  Proceedings of the IEEE International Conference on Computer Vision
  Workshops. pp. 397--403 (2013)

\bibitem{sari2019automatic}
Sar{\i}, C., Salah, A.A., Akdag~Salah, A.A.: Automatic detection and
  visualization of garment color in {Western} portrait paintings. Digital
  Scholarship in the Humanities  \textbf{34}(Supplement\_1),  i156--i171 (2019)

\bibitem{sheng2019generating}
Sheng, S., Moens, M.F.: Generating captions for images of ancient artworks. In:
  Proceedings of the 27th ACM International Conference on Multimedia. pp.
  2478--2486 (2019)

\bibitem{silva2021automatic}
Silva, J.M., Pratas, D., Antunes, R., Matos, S., Pinho, A.J.: Automatic
  analysis of artistic paintings using information-based measures. Pattern
  Recognition  \textbf{114},  107864 (2021)

\bibitem{sun2019deep}
Sun, K., Xiao, B., Liu, D., Wang, J.: Deep high-resolution representation
  learning for human pose estimation. In: Proceedings of the IEEE/CVF
  Conference on Computer Vision and Pattern Recognition. pp. 5693--5703 (2019)

\bibitem{wang2020computerized}
Wang, J.Z., Kandemir, B., Li, J.: Computerized analysis of paintings. In: The
  Routledge Companion to Digital Humanities and Art History, pp. 299--312.
  Routledge (2020)

\bibitem{yaniv2019face}
Yaniv, J., Newman, Y., Shamir, A.: The face of art: landmark detection and
  geometric style in portraits. ACM Transactions on graphics (TOG)
  \textbf{38}(4),  1--15 (2019)

\bibitem{zhang2016joint}
Zhang, K., Zhang, Z., Li, Z., Qiao, Y.: Joint face detection and alignment
  using multitask cascaded convolutional networks. IEEE Signal Processing
  Letters  \textbf{23}(10),  1499--1503 (2016)

\bibitem{zheng2020deep}
Zheng, C., Wu, W., Yang, T., Zhu, S., Chen, C., Liu, R., Shen, J., Kehtarnavaz,
  N., Shah, M.: Deep learning-based human pose estimation: A survey. arXiv
  preprint arXiv:2012.13392  (2020)

\end{thebibliography}
\end{document}